\documentclass[journal]{IEEEtran}
\usepackage{multirow}
\usepackage{graphicx}
\usepackage{subfigure}
\usepackage{amsmath}
\usepackage{amsfonts}
\usepackage{algorithm}
\usepackage{algorithmic}
\usepackage{makecell}
\usepackage{booktabs}
\setcellgapes{3pt}

\setcellgapes{3pt}

\usepackage[justification=centering]{caption}
\usepackage{color}
\usepackage{verbatim}
\usepackage{float}
\usepackage{url}

\usepackage{cite}

\ifCLASSINFOpdf
\else
\fi
\begin{document}
%
\title{Joint Optimization of Deployment and Trajectory in UAV and IRS-Assisted IoT Data Collection System}

\author{Li Dong, Zhibin Liu, Feibo Jiang and Kezhi Wang.
	\thanks{This work was supported in part by the National Natural Science Foundation of China under Grant nos. 41904127, 41604117, 62002115. in part by the Hunan Provincial Natural Science Foundation of China under Grant nos. 2020JJ4428, 2020JJ5105. in part by the Key Research and Development Plan of Hunan Province under Grant no 2021NK2020. (Corresponding author: Zhibin Liu)
		
	Li Dong (Dlj2017@hunnu.edu.cn) is with Changsha Social Laboratory of Artificial Intelligence, Hunan University of Technology and Business, Changsha, China,
    Zhibin Liu (lzb2000@hunnu.edu.cn) is with Hunan Provincial Key Laboratory of Intelligent Computing and Language Information Processing, Hunan Normal University, Changsha, China,	
	Feibo Jiang (jiangfb@hunnu.edu.cn) is with Hunan Provincial Key Laboratory of Intelligent Computing and Language Information Processing, Hunan Normal University, Changsha, China, 
    Kezhi	Wang (kezhi.wang@northumbria.ac.uk) is with the department of Computer and Information Sciences, Northumbria University}
}

\markboth{Submitted for Review}%
{Shell \MakeLowercase{\textit{et al.}}: Bare Demo of IEEEtran.cls for IEEE Journals}
%



\maketitle

\begin{abstract}

Unmanned aerial vehicles (UAV) can be applied in many Internet of Things (IoT) systems, e.g., smart farms, as a data collection platform.
However, the UAV-IoT wireless channels may be occasionally blocked by trees or high-rise buildings. Intelligent reflecting surface (IRS) can be applied to improve the wireless channel quality by smartly reflecting the signal via a large number of low-cost passive reflective elements.
This paper aims to minimize the energy consumption of the system by jointly optimizing the deployment and trajectory of the UAV. The problem can be formulated as a mixed-integer-and-nonlinear-programming (MINLP), which is difficult to be addressed by the traditional solution, which may be easily fall into the local optimal. To address this issue, we propose a Joint Optimization framework of depLoyment and Trajectory (JOLT), where an adaptive whale optimization algorithm (AWOA) is applied to optimize the deployment of the UAV, and an elastic ring self-organizing map (ERSOM) is introduced to optimize the trajectory of the UAV. Specifically, in AWOA, a variable-length population strategy is applied to find the optimal number of stop points, and a nonlinear parameter $a$ and a partial mutation rule are introduced to balance the exploration and exploitation. In ERSOM, a competitive neural network is also introduced to learn the trajectory of the UAV by competitive learning, and a ring structure is presented to avoid the trajectory intersection. Extensive experiments are carried out to show the effectiveness of the proposed JOLT framework.

\end{abstract}


%
\IEEEpeerreviewmaketitle

\section{Introduction}

	Unmanned aerial vehicles (UAV) can be applied in many Internet of Things (IoT) applications, e.g., smart farms\cite{6515307}, as a data collection platform, due to its feature of flexibility and easy to be deployed. 
	Additionally, as the UAV can move close to the IoT devices in the real environment, it can help reduce the energy consumption of IoT devices.
However, UAVs usually have stringent constraints of size, weight, and energy, which may limit their flight distance and time\cite{8907406}. Moreover, the line-of-sight (LoS) communication links may be occasionally blocked by some obstacles, e.g., buildings or trees.
To address the above-mentioned issues and improve the operation efficiency of the UAV system, 
intelligent reflecting surfaces (IRS) can be applied as a promising solution\cite{pan2020multicell} to help reflect and enhance the communication signal between UAV and the IoT devices. IRS is composed of a number of reflective elements, which can reflect the signal by adjusting their phase shift. IRS can be mounted on several places such as the walls/facades of buildings, which can significantly improve the quality of the communication links.

Based on the above background, we aim to optimize the UAV’s deployment and trajectory by minimizing the energy consumption of the whole system including the UAV and IoT devices. To achieve this goal, we propose a Joint Optimization framework of depLoyment and Trajectory (JOLT) which consists of an adaptive whale optimization algorithm (AWOA) and an elastic ring self-organizing map (ERSOM). The main contributions can be summarized as follows:

(1) The UAV and IRS-assisted IoT data collection system is proposed, where the UAV is introduced to collect the data and the IRS is applied to enhance the communication links between the UAV and the IoT devices. We formulate the optimization problem to minimize the energy consumption of the UAV and all the IoT devices by jointly optimizing the deployment and trajectory of the UAV.

(2) Then, the joint optimization framework named JOLT is proposed to solve the optimization problem efficiently, in which AWOA is presented to find the optimal deployment of the UAV, and ERSOM is applied to optimize the trajectory of the UAV.

(3) For the deployment design of the UAV, the optimal number of stop points is unknown and the problem is non-convex. Hence, a variable-length population strategy in the AWOA is presented to find the optimal number of stop points, and a nonlinear parameter $a$ and a partial mutation rule are introduced to balance the exploration and exploitation of the AWOA for searching the locations of the stop points. 

(4) For the trajectory planning of the UAV, ERSOM is applied as a competitive neural network which can learn the trajectory of the UAV by competitive learning between the neurons. We also introduce a ring structure in the ERSOM to avoid the trajectory intersection of the UAV.

The rest of our work is organized as follows. Section II surveys the related studies. 
The system model and problem formulation are introduced in Section III. Section IV describes the proposed JOLT framework. The simulation results and discussions are given in Section V. Finally, Section VI concludes the paper. 
\section{Related works}

\subsubsection{UAV deployment optimization} Wang \emph{et al.}\cite{8833501} optimized the location and number of UAVs through differential evolution algorithm with an elimination operator. Each individual in the population is represented as the location of UAV, and all population is represented as the deployment of UAV. Liu \emph{et al.}\cite{8720417} designed a genetic algorithm to optimize the deployment of UAVs. In the experiment, their algorithm was compared with the exhaustion search and the results showed that the proposed method could find a better solution with less computation. Reina \emph{et al.}\cite{REINA201816} designed an effective method to solve the multi-target coverage problem in the deployment of UAV, in which the optimization problem is based on the weighted fitness function. 

\subsubsection{UAV trajectory optimization} Yu \emph{et al.} \cite{YU2020106209} presented a new differential evolution algorithm for trajectory optimization, in which the choice of individuals is depended on the objective function and constraints. Li \emph{et al.} \cite{8943975} proposed a new trajectory optimization algorithm called MACO, which can decrease the probability of falling into the local optimum. Qu \emph{et al.} \cite{QU2020106099} applied a reinforcement learning based gray wolf optimization algorithm to solve the path planning of UAV. Shao \emph{et al.} \cite{SHAO2020415} proposed an improved particle swarm optimization algorithm to optimize the trajectory of UAV, in which the initial distribution of particles was improved by chaotic map. Yang \emph{et al.} \cite{yang2020optimal} introduced an optimal control strategy of winner-take-all model for target tracking and cooperative competition of multi-UAVs. Furthermore, Zuo \emph{et al.} \cite{zuo2022unmanned} summarized the flight control methods and future challenges of UAVs.

\subsubsection{IRS-assisted UAV system} Jiao \emph{et al.}\cite{9130430} designed  an IRS and UAV assisted multiple-input NOMA downlink network. Al-Jarrah \emph{et al.}\cite{9395180} analyzed the communication capacity of the IRS assisted UAV system, which was influenced by the imperfect phase information. Pan \emph{et al.}\cite{9367288} applied UAV and IRS to support terahertz (THz) communications by  optimizing trajectory of UAV, phase shift of IRS, terahertz subband allocation and power control jointly. You \emph{et al.} \cite{9690481} presented the promising application scenarios, issues, and potential solutions of jointly applying IRS and UAV in wireless networks.

However, the above works have not jointly optimized the deployment and trajectory of the UAV and the phase-shift matrix of the IRS to reduce the energy consumption of the UAV and all the IoT devices. Moreover, the computational complexity of the traditional solutions is high when the number of stop points is large. Hence, here we aim to design an efficient JOLT framework to optimize the deployment and trajectory of the UAV and the phase-shift matrix of the IRS jointly, where the number of stop points can be reduced and the computing time of the trajectory planning can be saved.

\section{System model and problem formulation}
\begin{figure}[htpb]
	\centering
	\includegraphics[width=8.8cm]{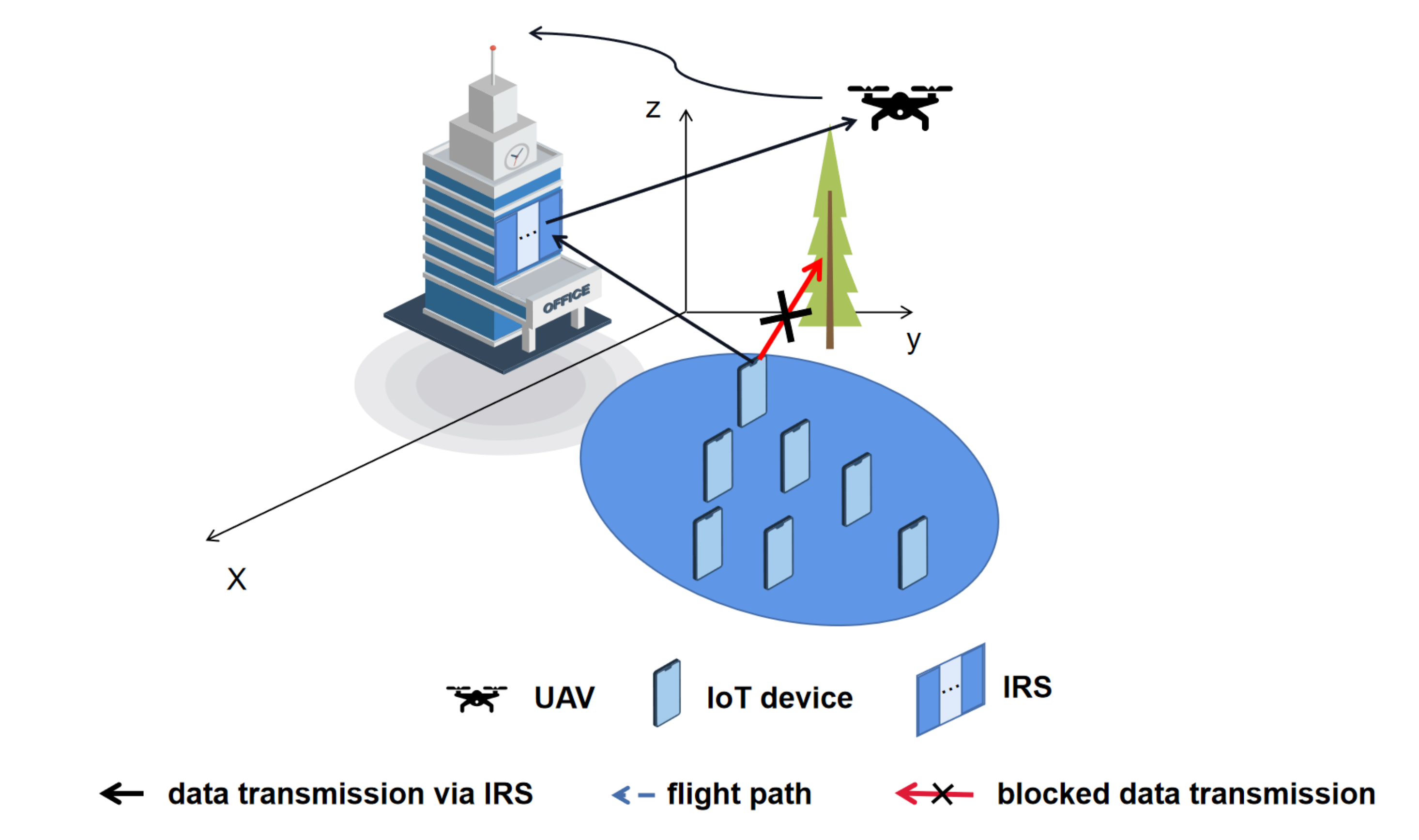}
	\caption{UAV and IRS-assisted IoT data collection system.}
	\label{fig:fig1}
\end{figure}
 In \textbf{Fig. \ref{fig:fig1}}, we design an IoT data collection system involving a UAV, an IRS and many IoT devices, where the set of IoT devices is $\mathcal{N}=\{1,2, \ldots, N\}$. The UAV can collect data through moving close to the IoT devices. If the LoS link is blocked, the transmission signals from IoT devices can be reflected and enhanced to the UAV via the IRS, which has an uniform linear array (ULA) with $M$ reflecting elements. 
Also, we assume that there are $k$ stop points and $k$ is a prior unknown, and we use $\mathcal{K}=\{1,2, \ldots, K\}$ to represent the set of the stop points for the UAV.

\subsection{Data transmission model}
We assume that the location of the $i$-th $(i \in \mathcal{N})$ IoT device is $\left(x_{i}^{D}, y_{i}^{D}\right)$, which denotes the coordinate of the $i$-th IoT device. The UAV flies at a fixed altitude $H$ and $q_j$ = $\left(X_{j}^{U}, Y_{j}^{U}, H\right)$ represents the coordinates of the $j$-th $(j \in \mathcal{K})$ stop point. Moreover, we consider that the coordinate of the IRS is represented by $\left(X^{I}, Y^{I}, H^{I}\right)$. Then, the distance between the $j$-th stop point and the IRS can be expressed as
    \begin{equation}\label{eq:Shi1}
    d_{j}^{U, I}=\sqrt{\left(X_{j}^{U}-X^{I}\right)^{2}+\left(Y_{j}^{U}-Y^{I}\right)^{2}+\left(H-H^{I}\right)^{2}}.
    \end{equation}

Similarly, the distance between the $i$-th IoT device and the IRS can be expressed as
	\begin{equation}\label{eq:Shi2}
	d_{i}^{I, D}=\sqrt{\left(X^{I}-x_{i}^{D}\right)^{2}+\left(Y^{I}-y_{i}^{ D}\right)^{2}+\left(H^{I}\right)^{2}}.
	\end{equation}
	
Then, the channel gain between the $j$-th stop point and the IRS can be expressed as \cite{wang2020joint}
	\begin{equation}\label{eq:Shi3}
     h_{j}^{U, I}
         =\sqrt{\frac{\alpha_1}{\left(d_{j}^{U, I}\right)^{2}}}\left[1, e^{-j \frac{2 \pi}{\lambda} d \phi_{j}^{U, I}}, \ldots, e^{\left.-j \frac{2 \pi}{\lambda}(M-1) d \phi_{j}^{U, I}\right]}\right]
	\end{equation}
\noindent where $\alpha_1$ is the path loss at 1m, the right of Eq. (\ref{eq:Shi3}) is the array response for the IRS\cite{tse_viswanath_2005}, where $\phi_{j}^{U, I}=\frac{X^{I}-X_{j}^{U}}{d_{j}^{U, I}}$  represents the cosine value of the arrival angle from the IRS to the $j$-th stop point of the UAV. $\lambda$ is the carrier wavelength, and $d$ is the antenna separation distance.

Similarly, the channel gain between the IRS and the $i$-th IoT device can be expressed as
	\begin{equation}\label{eq:Shi4}
	h_{i}^{I, D}=\sqrt{\frac{\alpha_2}{\left(d_{i}^{I, D}\right)^{2}}}\left[1, e^{-j \frac{2 \pi}{\lambda} d \phi_{i}^{I, D}}, \ldots, e^{-j \frac{2 \pi}{\lambda}(M-1) d \phi_{i}^{I, D}}\right]^{T}
	\end{equation}
where $\alpha_2$ is the path loss exponent, and $\phi_{i}^{I, D}=\frac{X^{I}-x_{i}}{d_{i}^{I, D}}$ is the cosine value of the angle of deviation from the IRS to the $i$-th IoT device. 

Additionally, we denote $\theta_{i, m, j} \in[0,2 \pi)$ $(m \in \mathcal{M})$ as the diagonal phase shift matrix. In the matrix, $i$ means the $i$-th IoT device, $m$ means the $m$-th reflecting element, and $j$ means the $j$-th stop point. Hence, the total matrix is $\Theta_{i, j}=\operatorname{diag}\left\{e^{\jmath \theta_{i, m, j}}, \forall m \in \mathcal{M} \right\}$.
Through the Eq. (\ref{eq:Shi3}) and (\ref{eq:Shi4}), we can get the transmission rate from the $j$-th stop point to the $i$-th IoT device, which can be represented as
	\begin{equation}\label{eq:Shi5}
   r_{ij}=\hat B \log _{2}\left(1+\frac{p_{i}\left|h_{j}^{U, I} \Theta_{i, j} h_{i}^{I, D}\right|^{2}}{\sigma^{2}}\right)
	\end{equation}
\noindent where $p_{i}$ is the transmit power. $\hat B$ is the system bandwidth. $\sigma^{2}$ is the  noise power. 
$a_{i j}$ is used to indicate whether the $i$-th IoT device sends data to the $j$-th stop point. To reduce power consumption, IoT devices should send data to the stop point with the fastest transmission rate. Therefore, $a_{i j}$ is given by
	\begin{equation}\label{eq:Shi6}
	\mathcal{C} 1: a_{i j}=\left\{\begin{array}{ll}
	1, & \text { if } j=\underset{j \in \mathcal{K}}{\arg\max }\,r_{i j} \\
	0, & \text { otherwise }
	\end{array}\right..
	\end{equation}
	
In addition, each IoT device can only be connected to one stop point, which can be expressed as
	\begin{equation}\label{eq:Shi7}
    \mathcal{C} 2: \sum_{j=1}^{K} a_{i j}=1,\ \forall i \in \mathcal{N}.
	\end{equation}
	
Additionally, considering the limitation of system bandwidth, each stop point is connected to no more than $\hat{M}$ IoT devices, which can be expressed as
	\begin{equation}\label{eq:Shi8}
	\mathcal{C} 3: \sum_{i=1}^{N} a_{i j} \leq \hat{M},\ \forall j \in \mathcal{K}.
	\end{equation}
	
For all data of IoT devices to be received, the condition listed below is required:
	\begin{equation}\label{eq:Shi9}
	\mathcal{C} 4: \sum_{i=1}^{N} \sum_{j=1}^{K} a_{i j}=N.
	\end{equation}
	
If the $i$-th IoT device has $D_{i}$ data sent to the UAV, the transmission time can be calculated as
	\begin{equation}\label{eq:Shi10}
	T_{i j}=\frac{D_{i}}{r_{i j}},\ \forall i \in \mathcal{N}, j \in \mathcal{K}.
	\end{equation}

Then, the energy consumption of the $i$-th IoT device can be calculated by the following equation:
	\begin{equation}\label{eq:Shi11}
	E_{i j}=p_{i} T_{i j}=\frac{p_{i} D_{i}}{r_{i j}},\ \forall i \in \mathcal{N}, j \in \mathcal{K}.
	\end{equation}

Hence, the energy consumption of all IoT devices is given by
	\begin{equation}\label{eq:Shi12}
    E_{i o t}=\sum_{i=1}^{N} \sum_{j=1}^{K} a_{i j} E_{i j}.
	\end{equation}
\subsection{UAV hovering model}
The data transmission rate is limited, so the UAV needs to hover at each stop point to collect all the IoT device data. Thus, the hover time of the UAV at the $j$-th stop point can be given by
	\begin{equation}\label{eq:Shi13}
	T_{j}^{H}=\max _{i \in \mathcal{N}}\left\{a_{i j} T_{i j}\right\},\ \forall j \in \mathcal{K}.
	\end{equation}

Then, the hovering energy consumption of the UAV at the $j$-th stop point can be represented as 
	\begin{equation}\label{eq:Shi14}
	E_{j}^{H}=p^{H} T_{j}^{H},\ \forall j \in \mathcal{K}
	\end{equation}
\noindent where $p^{H}$ denotes the hover power of the UAV.

Finally, the hovering energy consumption of the UAV can be given by
	\begin{equation}\label{eq:Shi15}
    E_{{hov}}=\sum_{j=1}^{K} E_{j}^{H}.
	\end{equation}
\subsection{UAV trajectory model}
After collecting data from the current stop point, the UAV will choose the next stop point, which can be denoted as $b_{j}^{i}$ ($ i,j \in \mathcal{K}$). Moreover, $b_{j}^{i}$ is equal to 1 if the UAV chooses the $j$-th stop points as the $i$-th point of the trajectory. Otherwise, $b_{j}^{i}$ is equal to 0. Since each stop point can be reached once, which is expressed as
\begin{equation}\label{eq:Shi16}
\sum_{i=1}^{K} b_{j}^{i}=1,\quad \forall j \in \mathcal{K}.
\end{equation}

Then, we define $q^{j}$ $(j \in \mathcal{K})$ as the location of the $j$-th point of the trajectory. Thus, the flight distance is give as
\begin{equation}\label{eq:Shi17}
	L=\sum_{j=1}^{K} \left\|q^{j}-q^{j-1}\right\|.
	\end{equation}

Finally, the flight energy consumption of the UAV can be expressed as
	\begin{equation}\label{eq:Shi18}
	E_{{fly}}=p^{F} L
	\end{equation}
where $p^{F}$ is the flight power of the UAV \cite{8663615}.

\subsection{Objective function and constraints}

The energy consumption of the UAV and IRS-assisted IoT data collection system includes the three items mentioned above. Therefore, the objective problem can be formulated as follows:
	\begin{equation}\label{eq:Shi19}
	\renewcommand{\arraystretch}{1.5}
	\begin{array}{l}
	
    \min\limits_{\left\{X_{j}^{U}, Y_{j}^{U}\right\},K,B,\Phi} E_{i o t}+\beta_1 E_{h o v}+\beta_2  E_{f l y}\\
\hspace{-0.65cm}\text{ s.t. }\mathcal{C} 1: a_{i j} \in\{0,1\},\ \forall i \in \mathcal{N}, j \in \mathcal{K}\\
	\mathcal{C} 2: \sum_{j=1}^{K} a_{i j}=1,\ \forall i \in \mathcal{N}\\
	\mathcal{C} 3: \sum_{i=1}^{N} a_{i j} \leq \hat{M},\ \forall j \in \mathcal{K} \\
	\mathcal{C} 4: \sum_{i=1}^{N} \sum_{j=1}^{N} a_{i j}=N \\
	\mathcal{C} 5: X_{\min }^{U} \leq X_{j}^{U} \leq X_{\max }^{U},\ \forall j \in \mathcal{K} \\
	\mathcal{C} 6: Y_{\min }^{U} \leq Y_{j}^{U} \leq Y_{\max }^{U},\ \forall j \in \mathcal{K} \\
	\mathcal{C} 7: k_{\min } \leq K \leq k_{\max } \\
	\mathcal{C} 8: \sum_{i=1}^{K} b_{j}^{i}=1, \forall j \in \mathcal{K}\\
	\mathcal{C} 9: \theta_{i, m, j} \in \Upsilon,  \forall i \in \mathcal{N},  j \in \mathcal{K}, m \in \mathcal{M}
	\end{array}
	\end{equation}
\noindent where $\beta_1, \beta_2 \geq 0$ are the weights factors; $X_{\min }^{U}$ and $X_{\max }^{U}$ are the
the minimum and maximum values of $X_{j}^{U}$, respectively; $Y_{\min }^{U}$ and $Y_{\max }^{U}$ are the minimum and maximum values of $Y_{j}^{U}$, respectively;  $k_{\min }$ and $k_{\max }$ are the minimum and maximum values of the optimal number of stop points $K$, respectively. $B=\left\{b_{i}^{j}, \forall i,j \in \mathcal{K} \right\}$ represents the trajectory of the UAV. $\Phi =\left\{\Theta_{i, j}, \forall i,j \in \mathcal{K}\right\}$ and $\Theta_{i, j}=\operatorname{diag}\left\{e^{\jmath \theta_{i, m, j}}, \forall m \in \mathcal{M} \right\}$. $\Phi$ is the phase-shift matrix of the IRS.
Due to the hardware limitation, the phase shift normally is set to some discrete values. Thus, similar to \cite{wang2020joint}, each element can only select $N^{I}$ phase shift values from set $\Upsilon = \left\{\frac{2 \pi}{N^{I}} i, i=0,1, \ldots, N^{I}-1\right\}$. Moreover, the UAV serves at most $\hat{M}$ IoT devices and at least one IoT device at each stop point.


\subsection{Solving the phase-shift matrix}
The quantitative passive beamforming method is applied to optimize the phase-shift matrix $\Phi$ of the IRS, in which Eq. (\ref{eq:Shi3}) can be rewritten as follows:
\begin{equation}\label{eq:Shi20}
\boldsymbol{h}_{j}^{U,I}=\left[\left|\boldsymbol{h}_{j}^{U,I}\right| e^{j \omega_{j, 1}^{U, I}},\left|\boldsymbol{h}_{j}^{U,I}\right| e^{j \omega_{j, 2}^{U, I}}, \ldots,\left|\boldsymbol{h}_{j}^{U,I}\right| e^{j \omega_{j, M}^{U, I}}\right]^{T}
\end{equation}
\noindent where $\left|\boldsymbol{h}_{j}^{U, I}\right|$ is the magnitude and $\omega_{j, m}^{U, I} \in[0,2 \pi)$ is the phase shift of the reflecting element $m$ in the IRS to the $j$-th stop point. Similarly, Eq. (\ref{eq:Shi4}) can be rewritten as follows:
\begin{equation}\label{eq:Shi21}
\boldsymbol{h}_{i}^{I, D}=\left[\left|\boldsymbol{h}_{i}^{I, D}\right| e^{j \omega_{i, 1}^{I, D}},\left|\boldsymbol{h}_{i}^{I, D}\right| e^{j \omega_{i, 2}^{I E}}, \ldots,\left|\boldsymbol{h}_{i}^{I, D}\right| e^{j \omega_{i, M}^{I, D}}\right]^{T}
\end{equation}
\noindent where $\left|\boldsymbol{h}_{i}^{I, D}\right|$ denotes the magnitude and $\omega_{i, m}^{I, D} \in[0,2 \pi)$ is the phase shift of the reflecting element $m$ in the IRS to the $i$-th IoT device.

In addition, the maximum received power can be coherently combined with signals from different paths, and this will maximize the data rate. Thus, we can optimize the phase shift $\theta_{i, m, j}$ of the $m$-th reflecting element in the IRS from the $i$-th IoT device to the $j$-th stop point with the following equation:
\begin{equation}\label{eq:Shi22}
\theta_{i, m, j}=\underset{\theta_{i, m, j}^{\prime} \in \Upsilon}{\operatorname{argmin}}\left|\theta_{j, i, m}^{\prime}-\left(\omega_{j, m}^{U,I}+\omega_{i, m}^{I,D}\right)\right| .
\end{equation}

\section{The proposed JOLT framework}
\subsection{Motivation and framework outline}

There are several heuristic search methods which may solve the proposed problems, e.g., \cite{huang2019non,wang2020multi,xu2020delaunay,meng2017new,liu2019some,meng2016approach}, but they all suffer from some common issues. We summarize the shortcomings of these heuristic search methods in solving the problem of UAV’s deployment and trajectory firstly, and then propose our joint optimization framework.

Firstly, the number of stop points is normally a preset value which is quite subjective in most existing methods. Therefore, the energy consumption of the whole system may not be minimized under the preset number. Moreover, the optimization of the deployment of UAVs is a non-linear optimization with multi-constraints, which is complicated and easy to fall into the local optimum. To address the above issues, we propose an AWOA algorithm to solve $\{X_{j}^{U}, Y_{j}^{U}\}$ and $K$ in Eq. (\ref{eq:Shi19}). In AWOA, the size of population is variable and the number of individuals is represented as the number of stop points, and each individual is denoted as the position of one stop point. At the same time, we also introduce two improvements, i.e., nonlinear parameter $a$ and partial mutation rule to enhance the search process of the AWOA.

Furthermore, for the trajectory optimization of the UAV, the traditional methods may not only converge to the local optimum easily, but also require a lot of computation time. In order to address the problem, we first use an ERSOM neural network to solve $B$ in Eq. (\ref{eq:Shi19}), where the competitive learning is used to map the neurons of the ERSOM to the trajectory of the UAV, and the ring structure is also introduced to the ERSOM to avoid the trajectory intersection.

The workflow of the JOLT framework is given in $\bf{Algorithm\enspace\ref{alg1}}$. Firstly, we  generate a population $\mathcal{P}$ as the initial positions of stop points randomly. During the iteration, the JOLT framework continues to generate the new population $Q$ by AWOA, and then we gradually adjust the number of stop points to find the optimal number. Specifically, in each iteration, we generate three new populations: $Q_{insert}$, $Q_{replace}$ and $Q_{delete}$. We can optimize the trajectory of each new population by ERSOM, and then we can calculate the objective function and obtain the total energy consumption for the whole system. If there is at least one population satisfies the constraints and achieve better performance than original $\mathcal{P}$, $\mathcal{P}$ will be updated. The JOLT framework will return better deployment and trajectory until the  maximum iteration number $T_{max}$ is reached.
\begin{algorithm}
	\caption{The JOLT framework}
	\label{alg1}
	\begin{algorithmic}[1]
		\STATE{Initialize the locations of stop points as $\mathcal{P}=\left\{({X_j}^U,{Y_j}^U), \forall j\in \mathcal{K}\right\}$ randomly until $\mathcal{P}$ satisfies the constraints in Eq. (\ref{eq:Shi19});}
		\WHILE{$t<T_{max}$}
		\STATE{Obtain $Q$ according to $\bf{Algorithm\enspace\ref{alg2}}$;}		
		
		\STATE{Generate $Q_{insert}$ by selecting an individual from $Q$ randomly and inserting it into $\mathcal{P}$;}
		\STATE{Generate $Q_{replace}$ by utilizing a randomly selected individual in $Q$ to replace the corresponding individual in $\mathcal{P}$;}
		\STATE{Generate $Q_{delete}$ by deleting an individual from $\mathcal{P}$ randomly;}
		
		\STATE{Evaluate whether $Q_{insert}$, $Q_{replace}$ and $Q_{delete}$ meet the constraints in Eq. (\ref{eq:Shi19});}
		\IF{at least one of the three new populations satisfies the constraints}
		\STATE{Calculate the phase-shift matrix $\Phi$ by Eqs. (\ref{eq:Shi20})-(\ref{eq:Shi22});}
		\STATE{Determine the trajectory $B$ of feasible population(s) by $\bf{Algorithm\enspace\ref{alg3}}$;}
		\STATE{Calculate the objective function according to $\{X_{j}^{U}, Y_{j}^{U}\}$, $\Phi$ and $B$;}
		\IF{the new population has better performance than $\mathcal{P}$}
		\STATE{The new population is used to replace $\mathcal{P}$;}
		\ENDIF
		\ENDIF
		\ENDWHILE\\
		\STATE{\textbf{return} $\mathcal{P}$, $\Phi$, $B$.}
	\end{algorithmic}
\end{algorithm}

\subsection{Adaptive whale optimization algorithm (AWOA)} \label{AWOA}
Whale optimization algorithm (WOA) is a fresh heuristic search proposed in \cite{MIRJALILI201651}, which consists of three parts inspired by the hunting behaviors of whales:

\subsubsection{Encircling prey} Whales can find and surround their prey, and this behavior is given by
	\begin{equation}\label{eq:Shi23}
	{X}(t+1)={X^{*}}(t)-{A} \cdot \left|{C} \cdot {X}^{*}(t)-{X}(t)\right|
	\end{equation}
\noindent where $t$ denotes the current number of iterations. ${A}$ and ${C}$ are coefficient vectors. $X^{*}$ is the location vector of the optimal solution currently obtained. ${X}$ is the current location vector. ${A}$ and ${C}$ are given by
\begin{equation}\label{eq:Shi24}
	{A}=2 {a} \cdot {r}-{a}
	\end{equation}
	\begin{equation}\label{eq:Shi25}
	{C}=2 \cdot {r}
	\end{equation}
\noindent where ${a}$ decreases from 2 to 0 linearly to control the exploration and exploitation phases during the search process. And ${r}$ is a random vector in [0,1].

\subsubsection{Bubble-net attacking} $|{A}|<1$ will cause the WOA to carry out local search. Specifically, whales follow a spiral path around their prey. Shrinking circle attacking is given by Eq. (\ref{eq:Shi23}) and spiral-shaped attacking is given by
\begin{equation}\label{eq:Shi26}
{X}(t+1)=\left|{X^{*}}(t)-{X}(t)\right| \cdot e^{b l} \cdot \cos (2 \pi l)+{X^{*}}(t)
\end{equation}
\noindent where $b$ is a constant for defining the shape of the logarithmic spiral. $l$ is a random number in [-1,1], and $\cdot$ is an element-by-element multiplication. Whales will randomly choose one of the two ways, i.e., encircling prey and bubble-net attacking to update the current location vector.

\subsubsection{Search for prey} $|{A}|>1$ will cause the WOA to carry out global search. The formula is given by 	
	\begin{equation}\label{eq:Shi27}
	{X}(t+1)={X}_{{rand}}(t)-{A} \cdot \left|{C} {X}_{{rand}}(t)-{X}(t)\right|
	\end{equation}
\noindent where ${X}_{{rand}}(t)$ is a random position vector (a random whale) chosen from the current population.

\begin{figure}[htpb]
	\centering
	\subfigure[The encoding mechanism of traditional WOA]{\includegraphics[width=8.8cm]{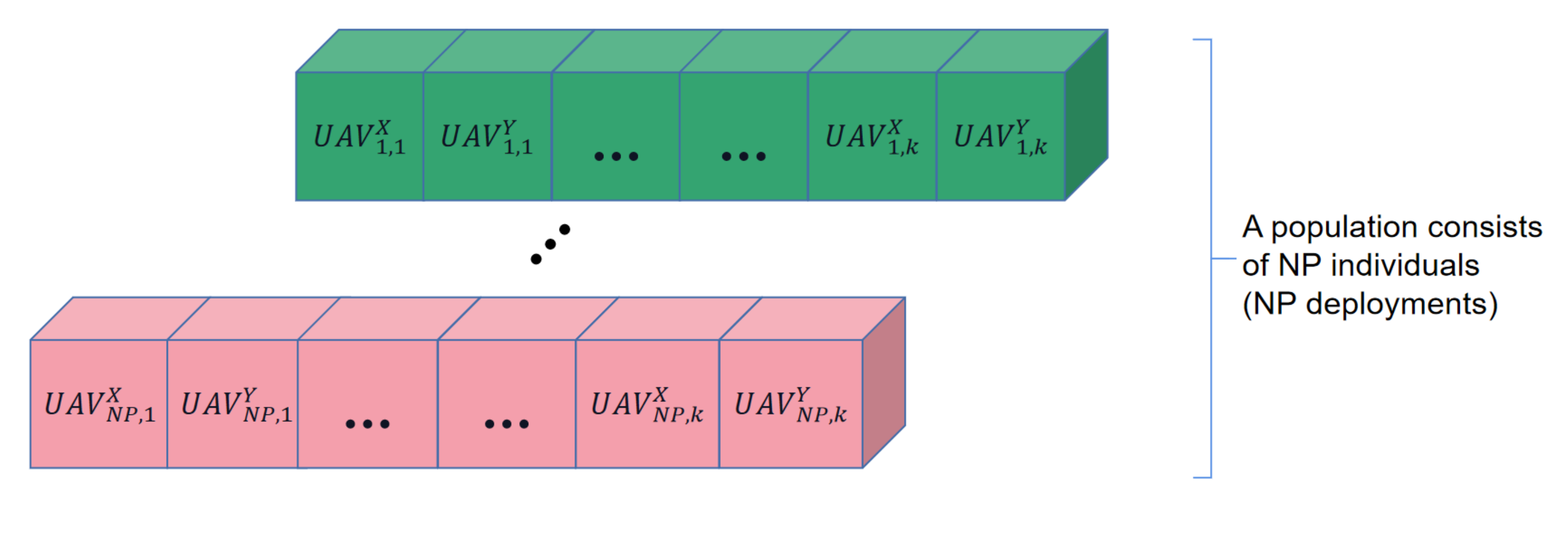}}
	\subfigure[The proposed encoding mechanism of the AWOA]{\includegraphics[width=8.8cm]{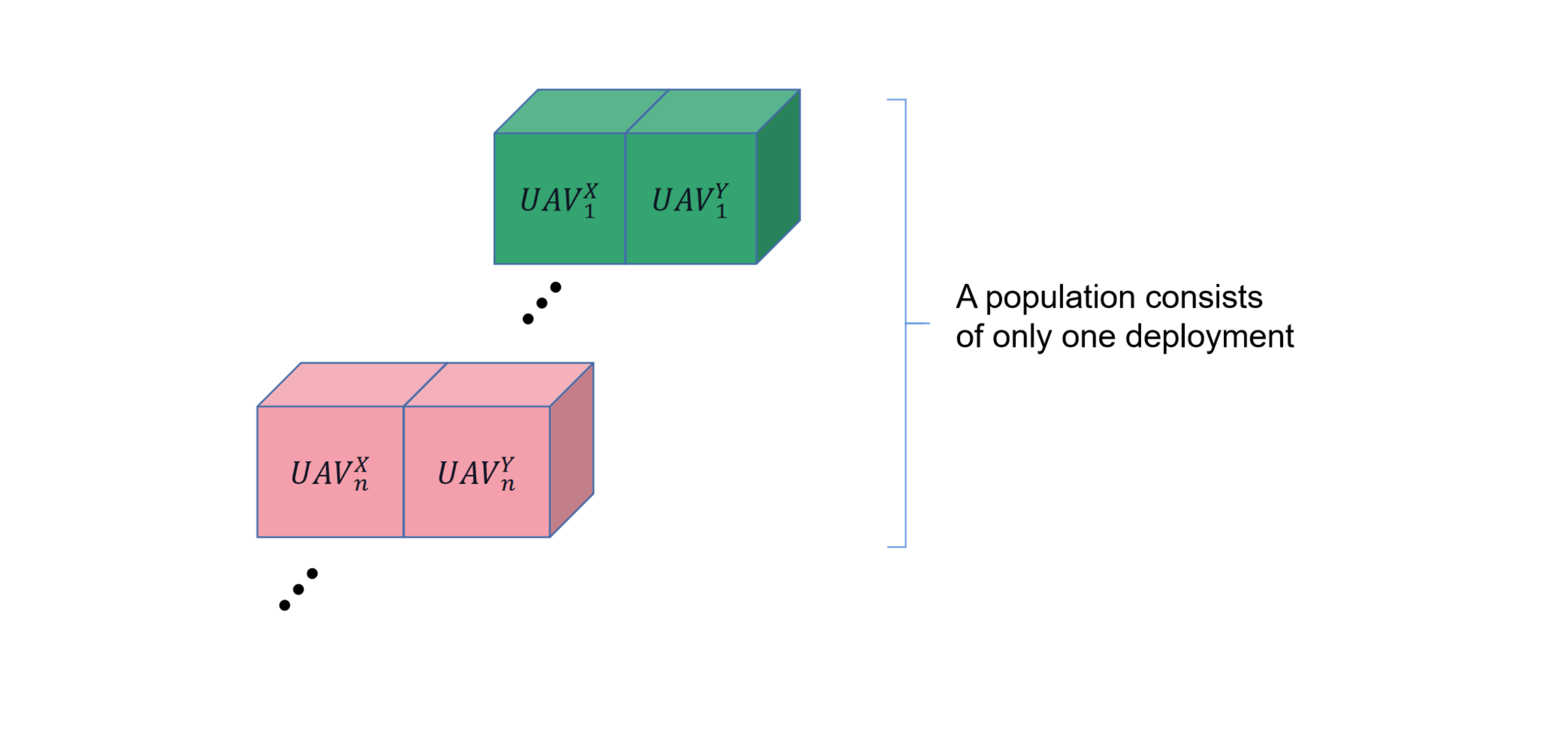}}
	\caption{Comparison of different encoding mechanisms}
	\label{fig:fig3}
\end{figure}

We aim to optimize the deployment of the UAV by WOA, which includes the optimal number of stop points and the positions of all stop points. However, the traditional heuristic search is hard to achieve this two targets concurrently\cite{8852863}. Next we propose the AWOA which improves the traditional WOA from three aspects:

\textbf{Variable-length population strategy}: 
 In the encoding mechanisms of traditional WOA, each individual represents the positions of all stop points and the whole population includes many feasible deployments. However, the length of individual is fixed which means the number of stop points should be preset. Inspired by \cite{8852863,8833501}, we improve the encoding mechanism of the WOA in which each individual represents only one position of the stop point and the whole population includes all stop points. by variable-length population strategy, we can optimize the number of stop points and the positions of stop points simultaneously by minimizing the total energy consumption. \textbf{Fig. \ref{fig:fig3}} shows two different encoding mechanisms.

 Moreover, in the proposed variable-length population strategy, $X^{*}$ is the optimal population, ${X}_{{rand}}$ is a random population, and $X$ is the current population. The update equation of AWOA can be redesigned as follow:
 \begin{small}
 \begin{equation}\label{eq:Shi28}
 	{X}(t+1)=\\
 	\left\{\begin{array}{ll}
 		{X^{*}}(t)-{A} \cdot \left|{C} \cdot {X}^{*}(t)-{X}(t)\right|, \text { if } p<0.5 \\
 		 \left|{C} \cdot {X^{*}}(t)-{X}(t)\right| \cdot e^{b l} \cdot \cos (2 \pi l)+{X^{*}}(t), \\\text { otherwise }
 	\end{array}\right.
 \end{equation}
\end{small}
\noindent where $p$ is a random number in [0,1].

\begin{figure}[htpb]
 		\centering
 		\includegraphics[width=8.8cm]{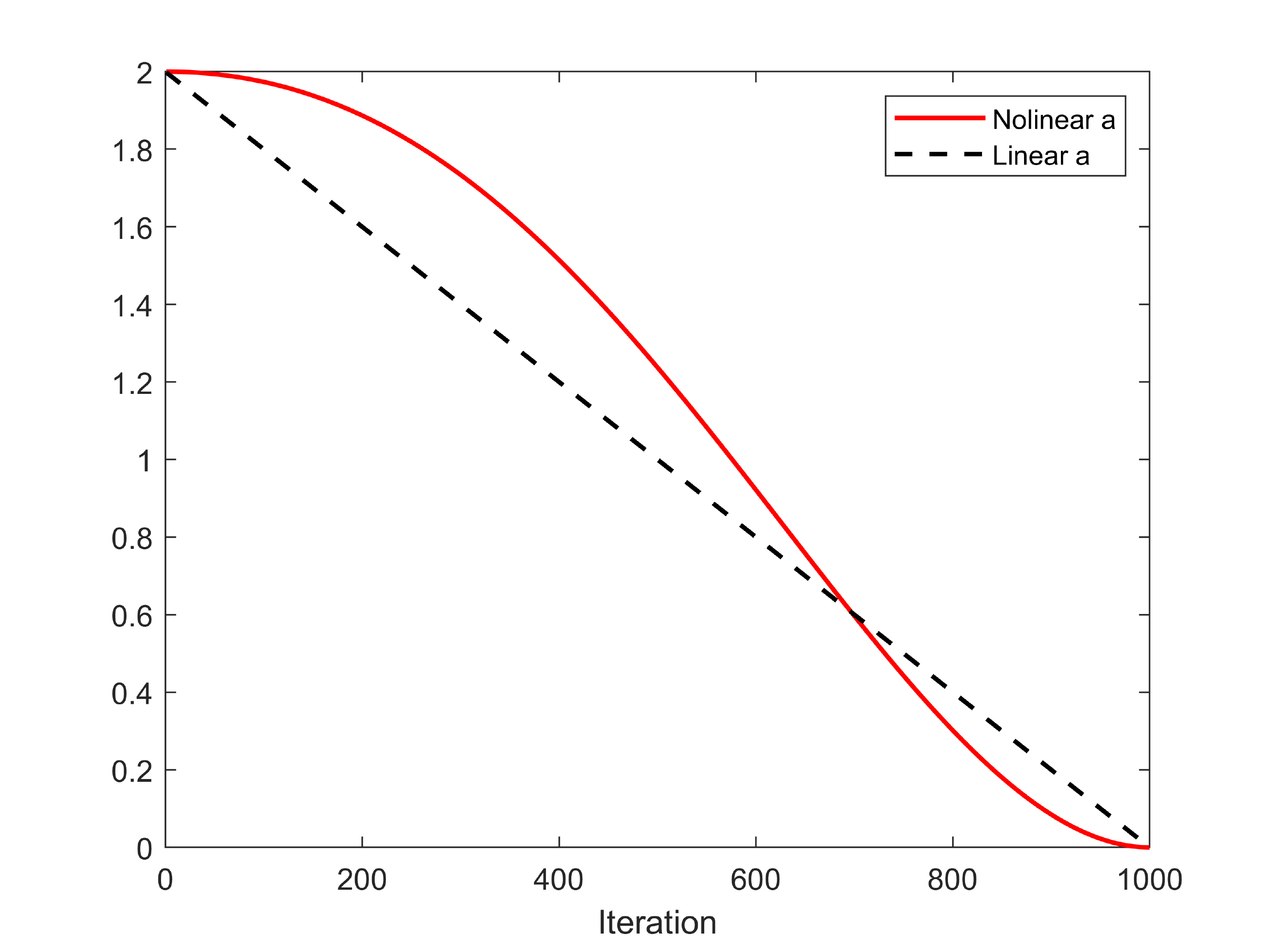}
 		\caption{Comparison of different $a$}
 		\label{fig:fig4}
 	\end{figure}

 	\textbf{Nonlinear parameter $a$}: ${a}$ is an important parameter in the WOA for balancing exploration and exploitation. The algorithm is inclined to global search when ${a}$ is large, while the algorithm is inclined to local search when ${a}$ is small \cite{9097175,9540904}. However, ${a}$ decreases linearly in traditional WOA, which results in an insufficient global search in the early iteration and slow convergence in the late iteration. Thus, we redesign a nonlinear parameter ${a}$ as follows:
 	
 	\begin{equation}\label{eq:Shi29}
 	{a}=\left(2-2 \cdot \frac{t^{3}}{T_{max }^{3}}\right) \cdot \cos \left(\frac{\pi}{2} \cdot \frac{t}{T_{max }}\right)
 	\end{equation}
 where $t$ indicates the current iteration and $T_{max}$ indicates the maximum iteration. At the beginning of the iteration, ${a}$ decreases slowly, then the search agent can make global search more sufficient. As iteration progresses, ${a}$ decreases distinctly and the algorithm converges rapidly. The trend of ${a}$ is shown in \textbf{Fig. \ref{fig:fig4}}. The red curve is calculated according to Eq. (\ref{eq:Shi29}), and the value of the black curve is reduced from the original ${a}$.

\textbf{Partial mutation rule}: 
Because of the variable-length population strategy, the population can only represent one deployment solution and the AWOA is easy to fall into the local optimum. Therefore, we design a partial mutation rule to avoid premature. If the search agent fails to find a better solution after a certain number of $\tau$ iterations, we think the algorithm has been trapped in a local optimal point, and then we will randomly select a part of the population to carry out the mutation rule as follows:

	\begin{equation}\label{eq:Shi30}
    	X_{rand}(t+1)=X_{rand}(t)+\rho \cdot  w_{c}(t+1)
    	\end{equation}
    \begin{equation}\label{eq:Shi31}
    	w_{c}(t+1)=\mu w_{c}(t)\left(1-w_{c}(t)\right)
    \end{equation}
\noindent where $X_{rand}(t)$ is the random individuals partially selected from the current population $X(t)$, $\rho$ is the scale parameter, and Eq. (\ref{eq:Shi31}) is a logistic equation, in which $w_{c}(t) \in(0,1)$ and $\mu$=4 is the chaotic control factor. 

After the optimization of the number and the coordinates of stop points by AWOA, we can calculate the diagonal phase-shift matrix $\Theta_{ij}$ and data rate $r_{ij}$ between different stop points and IoT devices according to Eq. (\ref{eq:Shi5}), and then calculate the data transmission energy consumption. The details of the AWOA are described in $\bf{Algorithm\enspace\ref{alg2}}$.

\begin{algorithm}
	\caption{AWOA}
	\label{alg2}
	\begin{algorithmic}[1]
	
		\STATE{Update ${a}$ according to Eq. (\ref{eq:Shi29});}
		\STATE{Update ${A}$ and  ${C}$ according to Eqs. (\ref{eq:Shi24})-(\ref{eq:Shi25});}
		\IF{$|{A}|>1$}
		\STATE{Generate $Q$ according to Eq. (\ref{eq:Shi27}); } 
	\ELSE
	\STATE{Generate $Q$ according to Eq. (\ref{eq:Shi28});}
		\ENDIF
	\IF{  $\mathcal{P}$ is not improved in $\tau$
		iterations}
\STATE{Update $Q$ according to Eqs. (\ref{eq:Shi30})-(\ref{eq:Shi31});}
\ENDIF
\STATE{\textbf{return} $Q$.}

	\end{algorithmic}
\end{algorithm}

\subsection{Elastic ring self-organizing map (ERSOM)} \label{ERSOM}
The ERSOM is a competitive neural network with a ring structure which includes a group of neural cells. The competitive learning is applied to find the nearest neural cell to a stop point each time, which is called the winner, and then moves the locations of the neighboring cells to the stop point by updating the weight vector of the selected neighboring cells. During the learning process, ERSOM will insert new neural cells gradually and the neural cells will learn the pattern behind the locations of all stop points and form a optimal trajectory. Then, we describe the procedure of the ERSOM.
\begin{figure}[htpb]
	\centering
	\includegraphics[width=8.8cm]{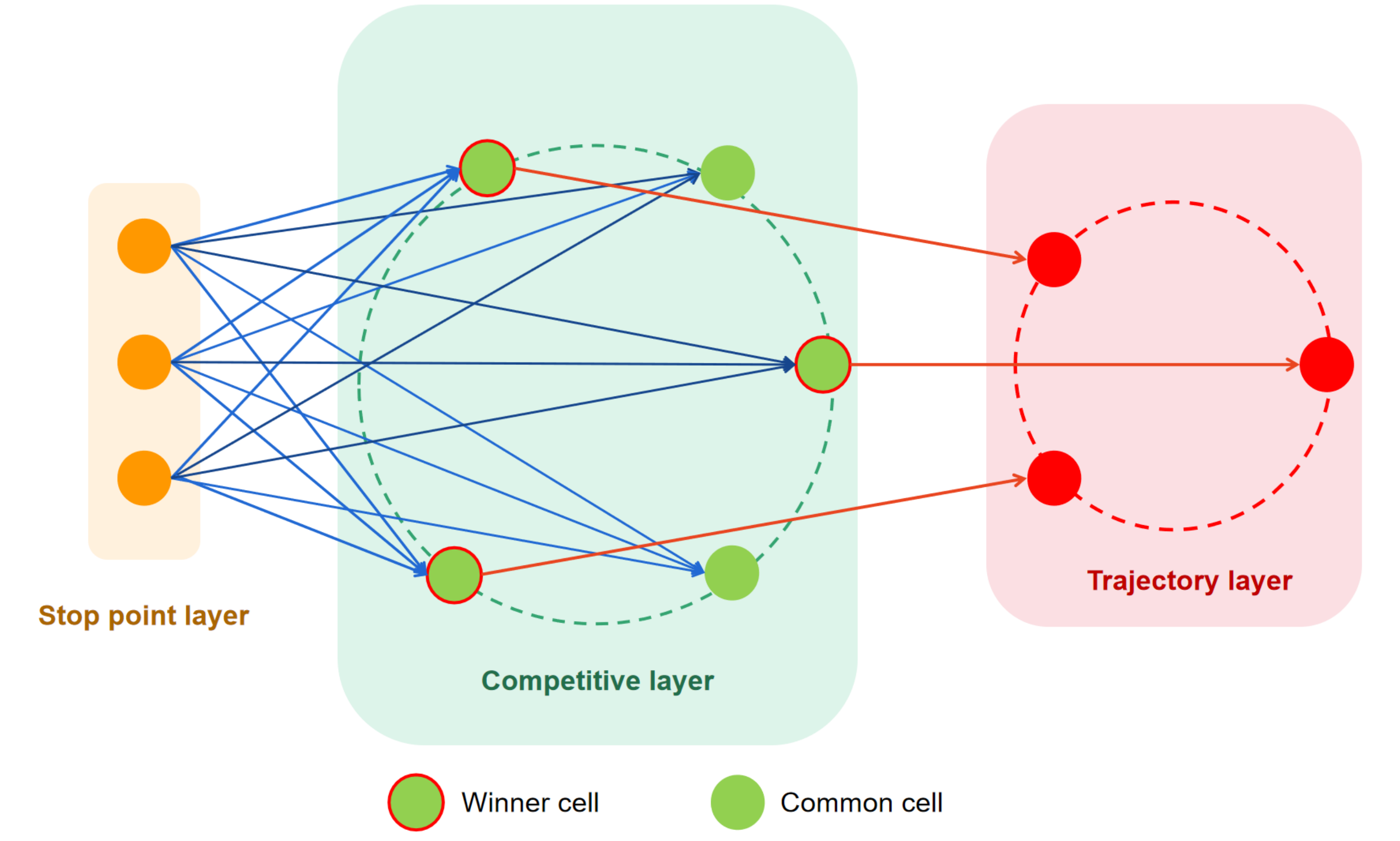}
	\caption{Structure of the ERSOM}
	\label{fig:fig5}
\end{figure}

\subsubsection{Construction of ERSOM} We can get a detailed description of the ERSOM from \textbf{Fig. \ref{fig:fig5}}, where the stop point layer includes the locations of all stop points. The trajectory layer and the competitive layer are both circular neural structures. The weight matrix between the stop point layer and the competitive layer records the spatial relationship between all stop points learned by the ERSOM. Finally, the trajectory layer outputs the optimal trajectory of the UAV by selecting all winners. The competitive layer of the ERSOM has a ring topology with $\hat{N}$ cells, denoted as $\mathcal{\hat{N}}=\{1,2,...,\hat{N}\}$, and each cell connects to two neighboring cells on both sides. Synaptic weight vector $w_{i}(t)$ can be denoted as the location of the $i$-th cell at the $t$-th iteration and a signal counter $C_i$ is used to record the winner history of the $i$-th cell. $\hat{N}(t)$ denotes the number of cells at the $t$-th iteration. At the beginning of the competitive learning, we initialize $\hat{N}(0)$, $C_{i}$ and $w_{i}(0)$ of the ERSOM, where $i \in \{1, \cdots, \hat{N}(0)\}$, and the locations of all stop points in the stop point layer can be obtained from the deployment of the UAV. $G_{max}$ is the maximum iteration number.

\subsubsection{Determination of winner} Cell $c$ will be selected as a winner at the $t$-th iteration if its synaptic vector $w_{c}(t)$ is closest to the $j$-th input stop point $X_j$, which can be expressed as
		\begin{equation}\label{eq:Shi32}
		\left\|X_j-w_{c}(t)\right\|=\min _{i}\left\|X_j-w_{i}(t)\right\|
		\end{equation}
\noindent where $\|\cdot\|$ denotes the Euclidean vector norm.
    
\subsubsection{Update of synaptic vectors and counters} Synaptic vectors of the winner $c$ and its neighbors will be updated, and the updating formula is given by
    	\begin{equation}\label{eq:Shi33}
    	w_{i}(t+1)=\left\{\begin{array}{ll}
    	w_{i}(t)+\beta\left(X_j-w_{i}(t)\right) & \text { if } i \in  \mathcal{\hat{N}}_{c} \\
    	w_{i}(t) & \text { otherwise }
    	\end{array}\right.
    	\end{equation}
\noindent where $\beta$ is the learning rate. $\mathcal{\hat{N}}_c=\{c-1, c, c+1\}$ denotes the neighbor set of the winner $c$. Then, the signal counter $C_w$ of the winner $c$ will add one and the other signal counters will preserve their original values at the $t$-th iteration.

\subsubsection{Insertion of neural cell} ERSOM will insert a new neural cell at every $T_r$ iterations so that the circular neural structure can learning more complex trajectory from the stop point layer. Meanwhile, $\hat{N}(t+1)=\hat{N}(t)+1$. We select one cell $p$ whose signal counter $C_{p}$ achieves the maximum value.
Moreover, if there are multiple maximum counter values, we will select one of them randomly. Then, we will select the neighbor $q$ who is closer to cell $q$ as follows
    	\begin{equation}\label{eq:Shi34}
        q=\left\{\begin{array}{ll}
        p-1 & \text { if }\left\|w_{p-1}-w_{p}\right\| \geq\left\|w_{p+1}-w_{p}\right\| \\
        p+1 & \text { otherwise }
        \end{array}\right.
    	\end{equation}

A new cell $r$ will be inserted between cell $p$ and $q$. $\mathcal{\hat{N}}_r=\{p, r, q\}$ denotes the neighbor set of the $r$-th cell. The synaptic vector of cell $r$ is initialized  as follows
\begin{equation}\label{eq:Shi35}
w_{r}=0.5\left(w_{p}+w_{q}\right).
\end{equation}


However, the more cells are inserted to the competitive layer, the more weights will be calculated and adjusted, and the computational efficiency of the ERSOM is lower. To solve the problem, we propose a novel elastic competitive layer, in which we can not only insert neural cells, but also remove neural cells. The adaptive deletion strategy is given as follows:

\textbf{Deletion of neural cell}:  
When the number of cells in the trajectory layer is more than twice the number of stop points in the stop point layer, we will select one cell $d$ to delete at every $\zeta T_r$ iterations, where $\zeta$ is a constant of deletion frequency and the signal counter $C_{d}$ has the minimum value as follows	
  \begin{equation}\label{eq:Shi36}
	C_d=\min C_i,\quad \forall i \in  \{1,2,...,\hat{N}\}
  \end{equation}
where $C_{d}$ is the signal counter of the cell $d$. The minimum value indicates that $d$ can not learn the pattern behind the locations of stop points sufficiently. By the deletion strategy, the number of cells will be restricted and the efficiency of the ERSOM will be improved. Additionally, the deletion strategy will not be implemented at the beginning of the competitive learning to ensure that there are a enough cells in the trajectory layer so that the patterns behind the locations of all stop points can be quickly learned.
    
\begin{figure}[htpb]
	\centering
	\subfigure[Initial phase]{
	\begin{minipage}[t]{0.2\textwidth}
		\centering
		{\includegraphics[width=1.8in]{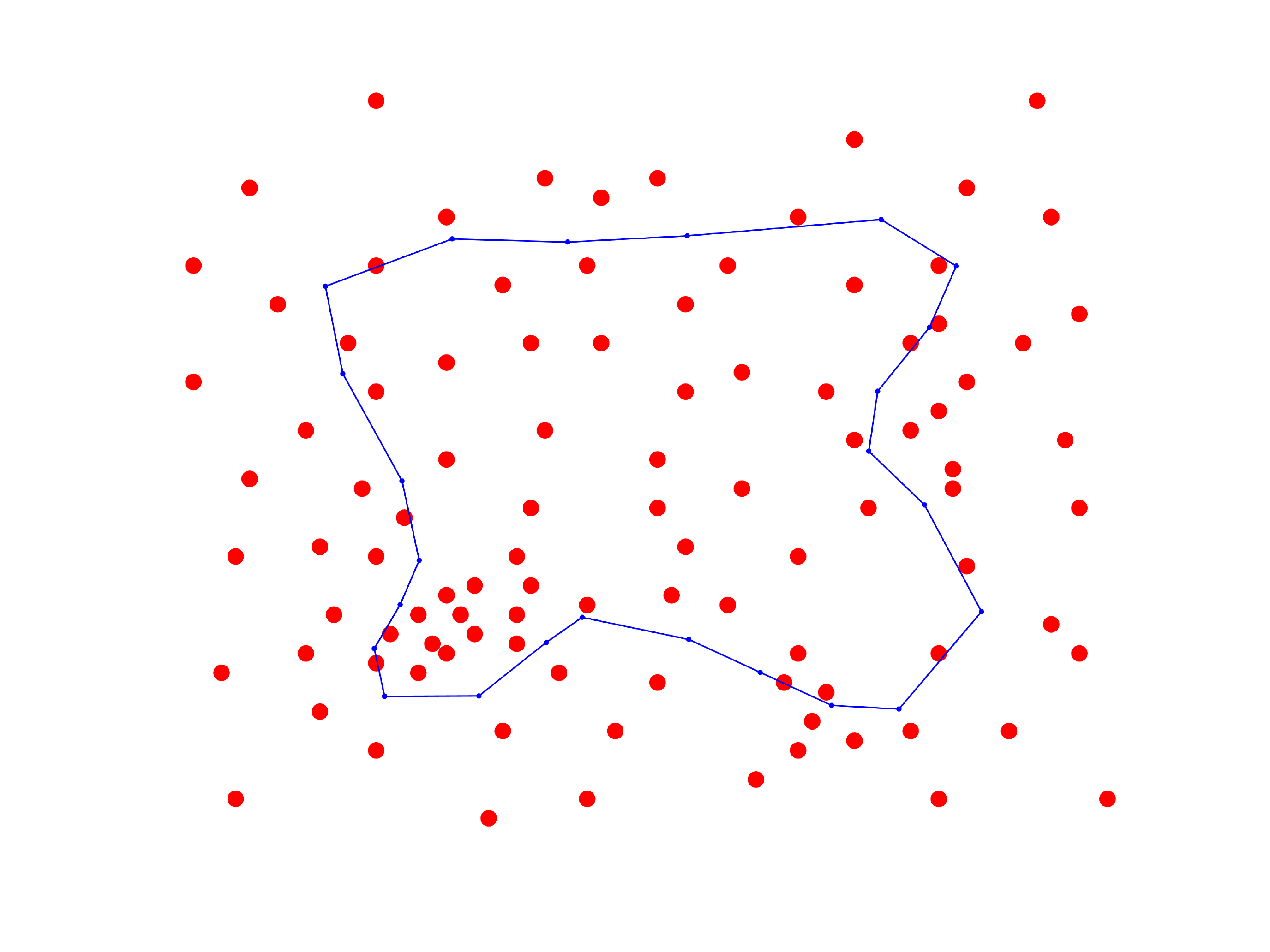}}
		\label{fig:figa}
	\end{minipage}}
\quad
	\subfigure[Learning phase 1]{\begin{minipage}[t]{0.2\textwidth}
			\centering
			{\includegraphics[width=1.8in]{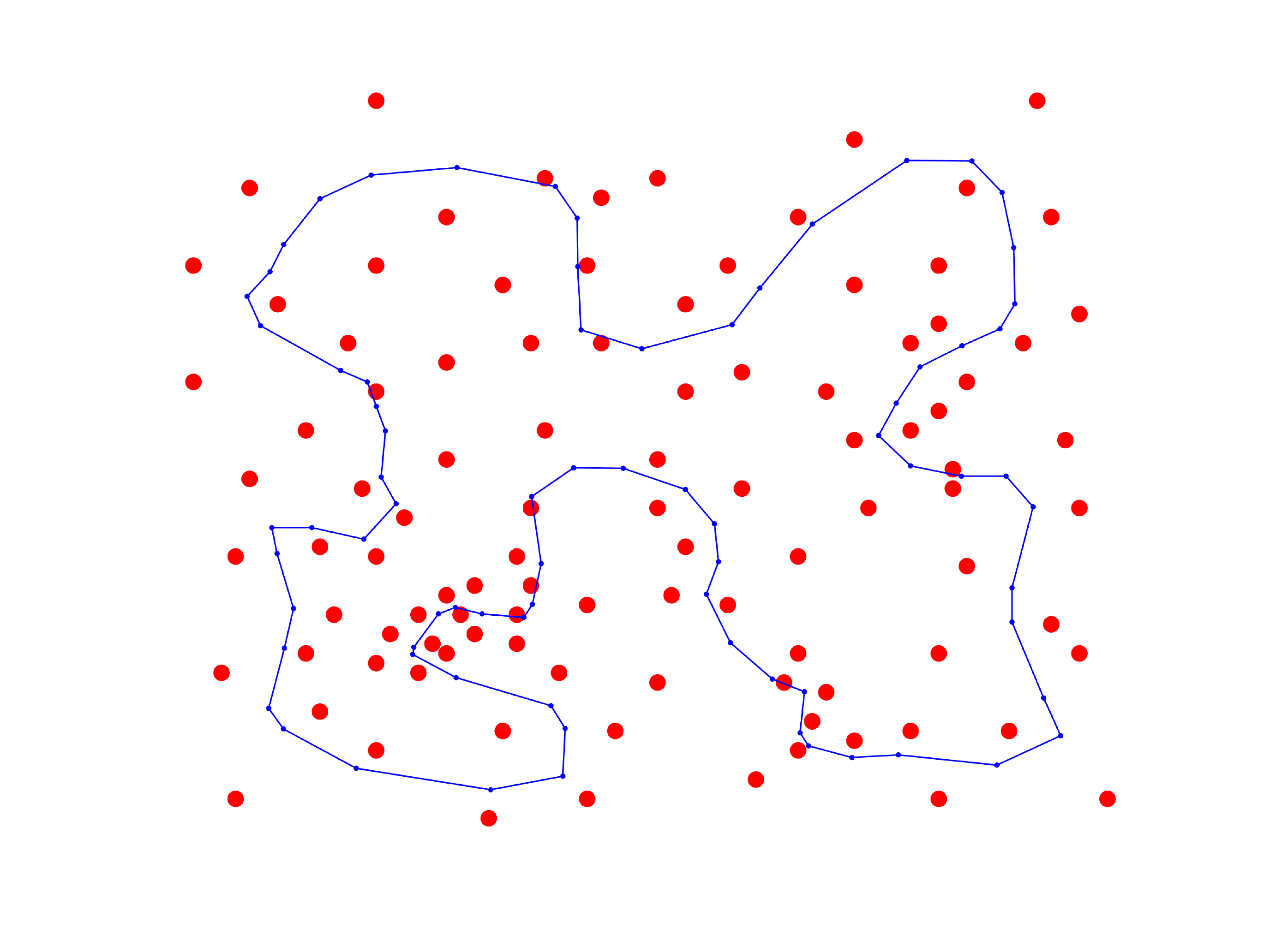}}
			\label{fig:figb}
	\end{minipage}}
\quad
	\subfigure[Learning phase 2]{\begin{minipage}[t]{0.2\textwidth}
			\centering
			{\includegraphics[width=1.8in]{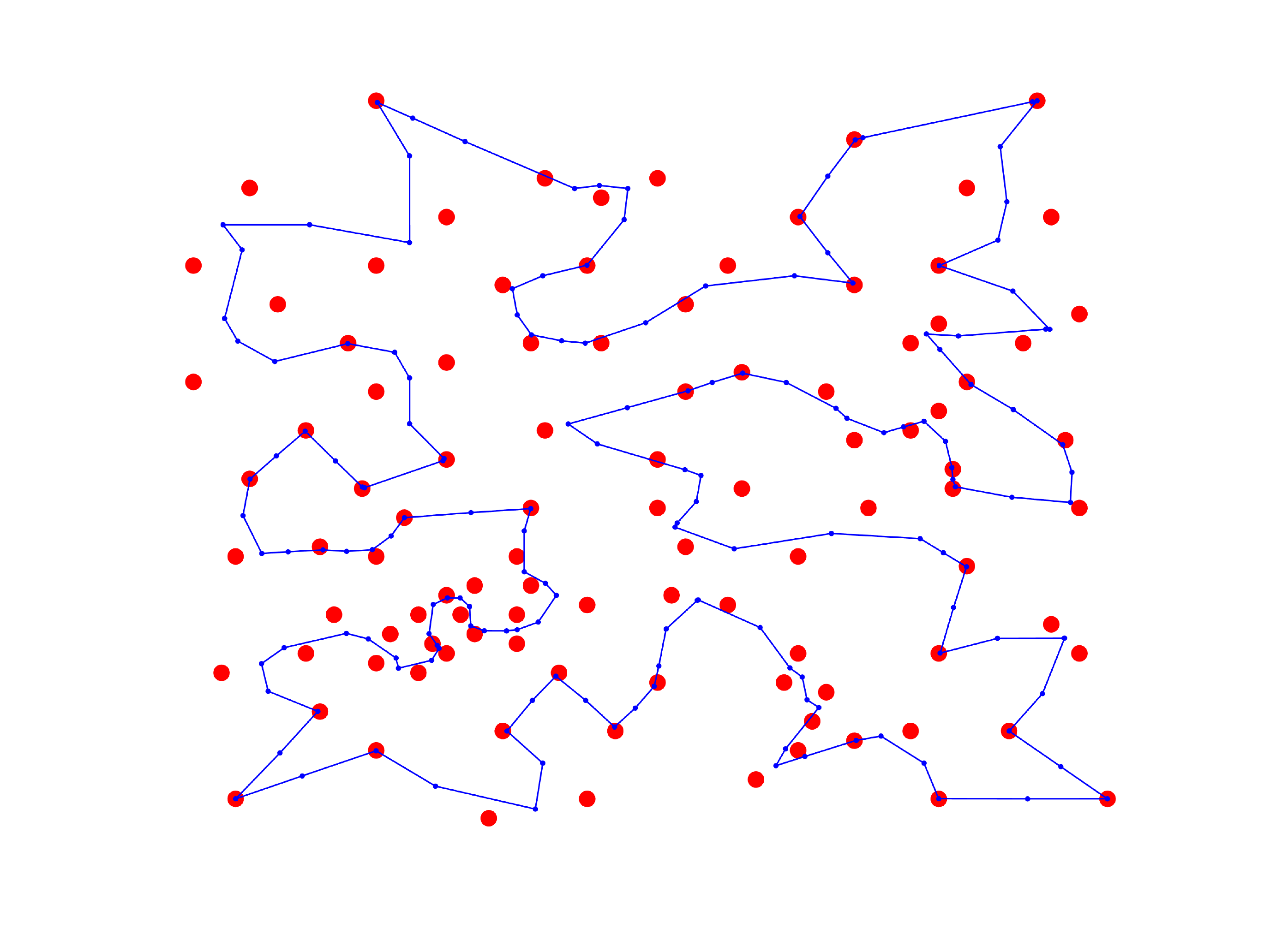}}
		\label{fig:figc}
	\end{minipage}}
\quad
	\subfigure[Final phase]{\begin{minipage}[t]{0.2\textwidth}
			\centering
			{\includegraphics[width=1.8in]{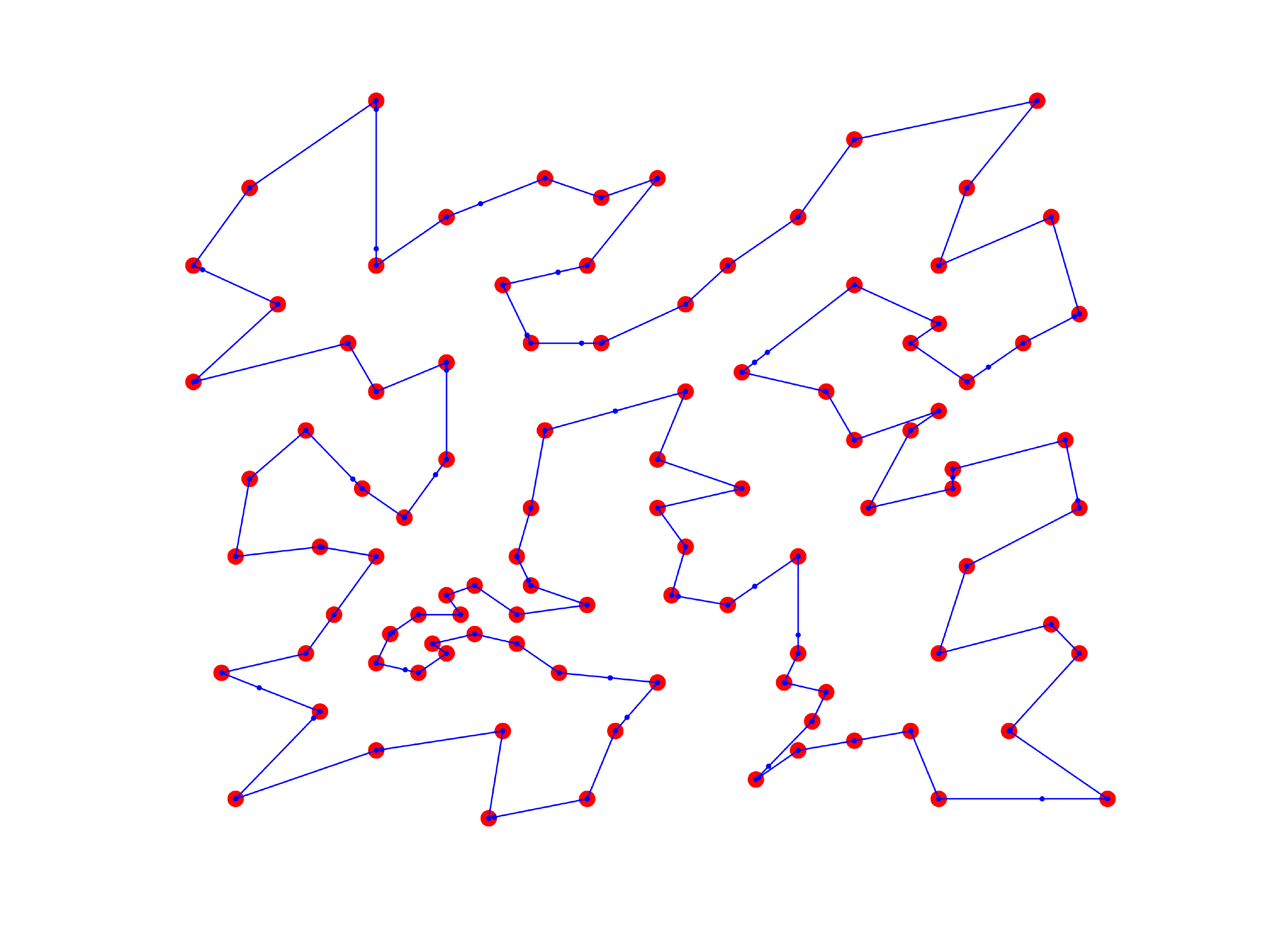}}
			\label{fig:figd}
	\end{minipage}}

	\caption{The learning process of the ERSOM.}
	\label{fig:fig6}
\end{figure}

\textbf{Generation of trajectory}: During the competitive learning, the number of neural cells will grow elastically and each neural cell will move to the nearest stop point. Finally, we select the nearest cell from each stop point. Due to the elastic ring topology of the ERSOM, each stop point can find a nearest cell, which will determine the ring trajectory. \textbf{Fig. \ref{fig:fig6}} shows the learning process of neural cells at the competitive layer, and $\bf{Algorithm\enspace\ref{alg3}}$ describes the details of the ERSOM. In \textbf{Fig. \ref{fig:fig6}}, the red dot is the stop point, the blue dot is the neural cell, and the blue line is the output trajectory. The flight path of the initial time is shown in the \textbf{Fig. \ref{fig:figa}}. During the learning process, new cells will be inserted and some unimportant cells will be deleted (e.g., \textbf{Fig. \ref{fig:figb} and \textbf{Fig. \ref{fig:figc}}}). Finally, the blue neural cells approach the red stop points and form a circular flight route for the UAV in \textbf{Fig. \ref{fig:figd}}.

 \begin{algorithm}
 	\caption{ERSOM}
 	\label{alg3}
 	\begin{algorithmic}[1]
 		\STATE{Initialize $w_i(0)$, $\hat{N}(0)$  and $C_i$;}
 	\WHILE{$t<G_{max}$}
 		\FOR{$j=1:K$}
 		\STATE{Choose the winner $c$ according to Eq. (\ref{eq:Shi32});}
 		\STATE{Update $w_{i}(t)$ according to Eq. (\ref{eq:Shi33});}
 		\STATE{Update $C_{w}=C_{w}+1$;}
 \ENDFOR\\
 		\IF{$t$ mod $T_r$= 0}
 		\STATE{Insert a new cell $r$ by Eq. (\ref{eq:Shi34})-(\ref{eq:Shi35});}
 		\STATE{$\hat{N}(t+1)=\hat{N}(t)+1$;}
 		\ENDIF
 		\IF{$t$ mod $\zeta T_r$= 0 and $\hat{N}(t+1)>2K$}
 		\STATE{Delete cell $d$ according to Eq. (\ref{eq:Shi36});}
 		\STATE{$\hat{N}(t+1)=\hat{N}(t)-1$;}
 		\ENDIF
 	\ENDWHILE\\
 	
 \STATE{Select the closest cell from each stop point and generate the trajectory $B$;}
 
	\STATE{\textbf{return}  $B$;}
 		
 	\end{algorithmic}
 \end{algorithm}

\subsection{Time complexity analysis}
	
The JOLT framework is composed of AWOA and ERSOM algorithms. The time complexity analysis is described as follows:
\begin{itemize}
	\item Time complexity of AWOA is $O(I_{woa} \times T_{max})$, where $T_{max}$ is the maximum iteration number of the AWOA, and $I_{woa}$ indicates the time complexity of the update of one feasible solution, which depends on the evaluation complexity of the feasible solution.
	\item Time complexity of ERSOM is $O(K \times G_{max})$, where $K$ is the number of stop points, and $G_{max}$ is the maximum iteration number of the ERSOM.
\end{itemize}

In the JOLT framework, the ERSOM is called to evaluate the feasible solution in each AWOA iteration. Hence, the time complexity of the JOLT framework is $O(K  \times G_{max} \times T_{max})$.

\section{Simulation results and discussions}

\subsection{Parameter settings}

In our simulations, all IoT devices are randomly placed in the areas with squared size $1000 \mathrm{m} \times 1000 \mathrm{m}$ and the IRS is placed at the center. The data size $D_i$ collected from IoT devices is randomly distributed within [1, $10^3$]MB. We use only one UAV to collect all data of IoT devices, and other parameters of the system model used in the simulations are summarized in \textbf{Table \ref{tabel1}}.  Moreover, all simulations are carried out in Matlab 2020b environment running on Intel Core i5-8250U CPU with 8 GB RAM.

\begin{table}[]
	\centering
	\makegapedcells
	\caption{Simulation parameters}
	\begin{tabular}{|p{150pt}|p{50pt}|}
		\hline
		Simulation parameters                                     & Value          \\ 
		\hline
		Flight height of the UAV                                      & 200m           \\
		\hline
		Maximum number of served devices  $\hat{M}$   & 5              \\
		\hline
		Noise power $\sigma^{2}$                                          & -250dBm         \\
		\hline
		Transmitting power $p_{i}$ & 0.1W           \\
		\hline
		Hover power $p^{H}$                               & 1000W          \\
		\hline
		Flight power $p^{F}$                               & 1283W          \\
		\hline
						Path loss exponent $\alpha_1$  and $\alpha_2$                                     & 0.01         \\
		\hline
		Weights factor $\beta_1$  and $\beta_2$                                     & 10000 and 0.5         \\
		\hline

		Bandwidth $\hat{B}$                                     & 1MHz           \\
		\hline
		
	\end{tabular}
	\label{tabel1}
\end{table}

\subsection{Performance evaluation of AWOA}

We evaluate the performance of the proposed AWOA, standard WOA \cite{MIRJALILI201651}, and nonlinear sine whale optimization algorithm (NSWOA)\cite{math6110220} in optimizing the deployment of the UAV. NSWOA is a variant of the standard WOA using a nonlinear $a$. In all experiments, the total energy consumption is defined as the fitness function of WOAs, and each algorithm is executed 10 times and the average energy consumption is calculated and recorded. Here are the parameter settings for different WOAs. 

\begin{enumerate}
	\item AWOA: The initial population size is set to 100; $b$ is set to 1; $\tau$ is set to 30; $\rho$ and $\mu$ are set to 1 and 4, respectively.
	\item NSWOA and WOA: The initial population population size is set to 100; $b$ is set to 1.
\end{enumerate}
 
We record the energy consumption (EC) and the computation time (CT) in \textbf{Table \ref{table3}}. It can be seen that the WOAs with nonlinear $a$ (e.g., NSWOA and AWOA) have lower EC than the standard WOA. It is because the nonlinear $a$ can balance the global search and local search effectively. In especial, AWOA achieves the lowest EC. The reason of the best performance can be explained by the partial mutation rule which can 
effectively jump out of the local extremum and find the global best solution. 

Then, we evaluate the computation time of different WOAs. In \textbf{Table \ref{table3}}, it can be seen that AWOA takes slightly longer computation time than other WOAs. This is because the extra partial mutation rule is introduced to the AWOA and the AWOA is more complex than other algorithms. However, this complexity can be compensated by its stable global search ability and fast convergence speed, and the AWOA achieves the lowest EC. This gives AWOA an advantage in some time-insensitive scenarios. For example, the IoT system may need to collect data from IoT devices at regular intervals, the deployment of the UAV can be planned before the data collection phase.

\begin{table}[]
	\footnotesize
	\caption{Comparison of energy consumption and computation time.}
	\renewcommand{\arraystretch}{1.5}
	\setlength{\tabcolsep}{14mm}{
\begin{tabular}{@{}lll@{}}
	\bottomrule
Algorithm	& EC & CT(s) \\ \midrule
AWOA	& 129.7827e4 & 280.61 \\ 
NSWOA	& 142.1726e4 & 272.07 \\ 
WOA	& 161.5351e4 & 252.46 \\ 
	\bottomrule
\end{tabular} }
	\label{table3}
\end{table}
    \subsection{Performance evaluation of ERSOM}\label{V.C}
   
\begin{figure}[htpb]
   	\centering
   	\subfigure[The early phase of the ERSOM learning process.]{\includegraphics[width=9.2cm]{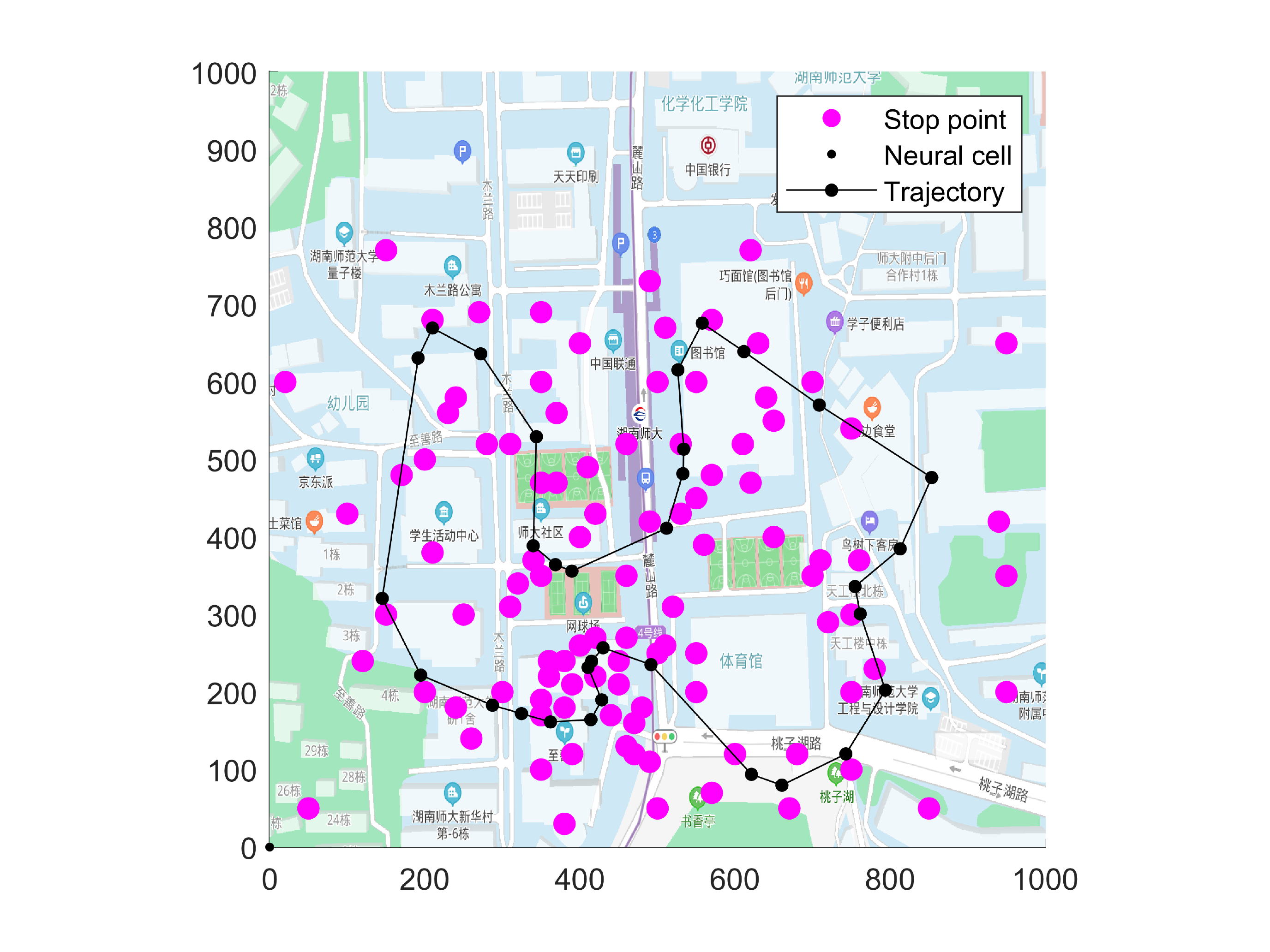}}
   	\subfigure[The final phase of the ERSOM learning process.]{\includegraphics[width=9.2cm]{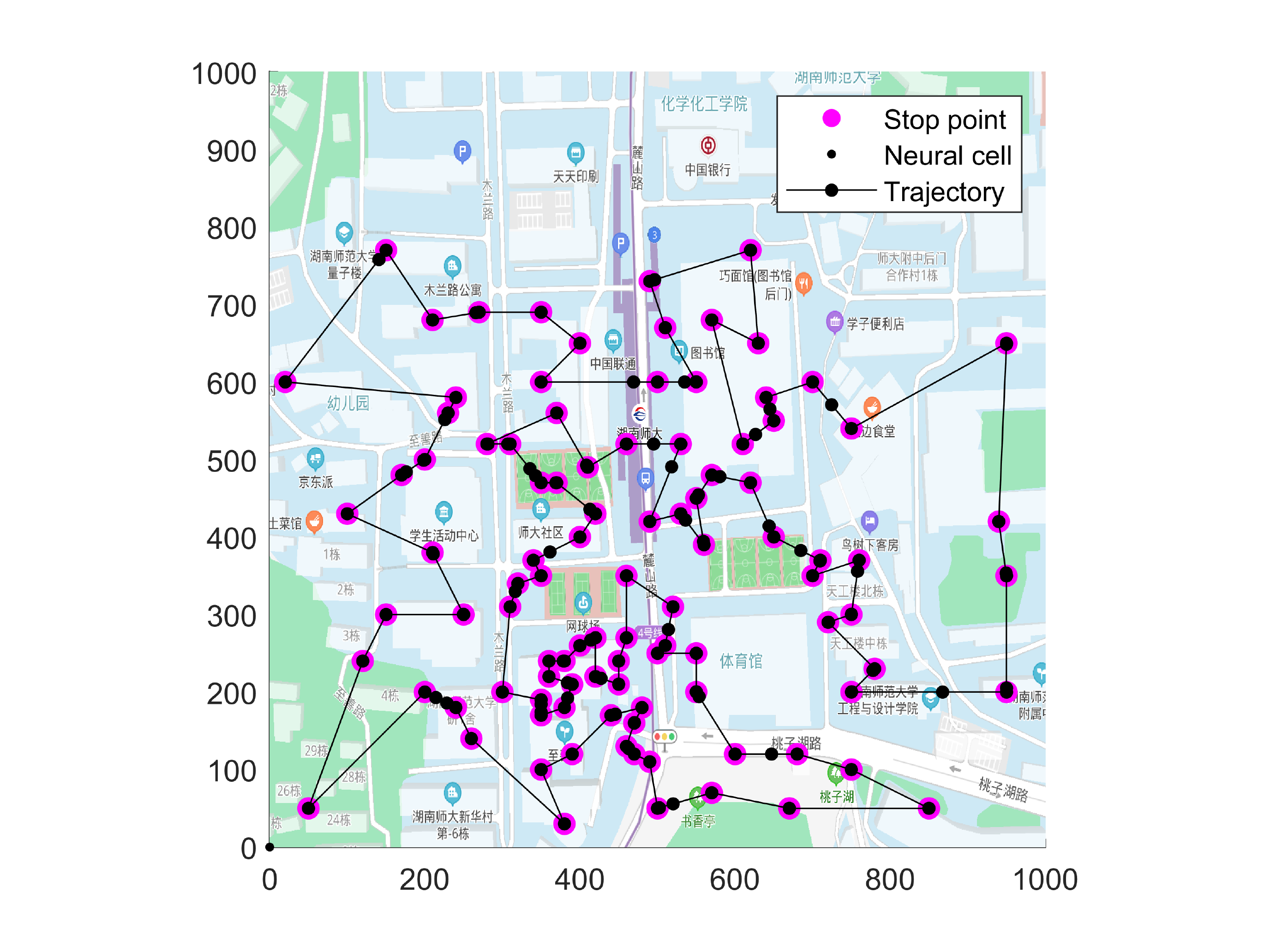}}
   	\caption{Performance evaluation of ERSOM.}
   	\label{fig:fig8}
   \end{figure}

    We evaluate the performance of ERSOM in trajectory optimization. Firstly, we consider a field case and visualize the optimization process of the trajectory design. The simulation scenario was selected at Hunan normal university in Hunan province of PR China, where a UAV is introduced to collect data form IoT devices within the range of 1000 meters by 1000 meters. The number of stop points is 101. The results are shown in \textbf{Fig. \ref{fig:fig8}}. The pink point represents the stop points, the black point represents the neural cells and the black line represents the trajectory of the UAV respectively. It can be seen that ERSOM algorithm can generate the optimal trajectory of the UAV efficiently.
    
    Next, we evaluate the performance of the proposed ERSOM, standard ring self-organizing map (RSOM)\cite{1202828} and DEVIPS\cite{8852863} on three different traveling salesman problem (TSP) data sets: att48, eil101 and tsp225 \cite{choong2019artificial}. RSOM is a neural network-based solver and DEVIPS is a heuristic search-based solver for large scale TSP. Here are the parameter settings for different algorithms.

    	\begin{enumerate}
    		\item ERSOM: $\beta$ is set to 0.1; $T_{\text {r}}$ is set to 5; $\zeta$ is set to 3.
    		\item RSOM: $\beta$ is set to 0.1; $T_{\text {r}}$ is set to 5.
    		\item DEVIPS: The initial population size is set to 100; The scaling factor $F$ is set to 0.6, and the crossover parameter $CR$ is set to 0.5.
\end{enumerate}

   We compare the ERSOM with other contenders on three classic tsp data sets. The numbers of stop points in data set att48, data set eil101 and data set tsp225 are 48, 101 and 225, respectively. We show the mean distance (Mean), CT of the three algorithms in \textbf{Table \ref{table6}}, and we also calculate the relative error (RE) based on the optimal distances (Optimum). It can be observed that the ERSOM achieves the lowest distance in att48 and eil101, and the ERSOM takes the least time in all TSP data sets. The high accuracy and high efficiency of the ERSOM can be explained by two reasons: on one hand, the neural network-based TSP solvers (e.g., RSOM and ERSOM) can find a sub-optimal solution of the TSP by competitive learning more quickly than the heuristic search. On the other hand, the adaptive deletion strategy is introduced to the ERSOM, so that the compact structure of the neural network is achieved and the computation time is reduced.
   
\begin{table}[]
	\caption{Comparison of different TSP solvers.}
	\renewcommand{\arraystretch}{1.3}
	\setlength{\tabcolsep}{3mm}{
		\begin{tabular}{@{}llllll@{}}
			\toprule
			Data set & Optimum & TSP solver & Mean & RE & CT(s) \\ \midrule
			&                     & ERSOM     & 35057.48   & 4.58\%        & 33.12      \\
			att48   & 33523.71         & RSOM      & 35297.35   & 5.29\%        & 37.23      \\
			&                     & DEVIPS        & 49247.00         & 46.90\%       & 43.62    \\
			\cline{1-6}
			&                     & ERSOM     & 677.00           & 5.40\%        & 54.43      \\
			eil101  & 642.30               & RSOM      & 683.17     & 6.36\%        & 58.96      \\
			&                     & DEVIPS        & 2083.97       & 224.45\%      & 61.77      \\
			\cline{1-6}
			&                     & ERSOM     & 4053.00          & 5.03\%        & 120.32     \\
			tsp225  & 3859.00                & RSOM      & 4045.00          & 4.82\%        & 138.54     \\
			&                     & DEVIPS        & 32774.27      & 749.29\%      & 535.87 \\ \bottomrule
	\end{tabular}}
	\label{table6}
\end{table}

\begin{figure}[htpb]
	\centering
	\subfigure[Number of IoT devices is 100.]{\includegraphics[width=8.8cm]{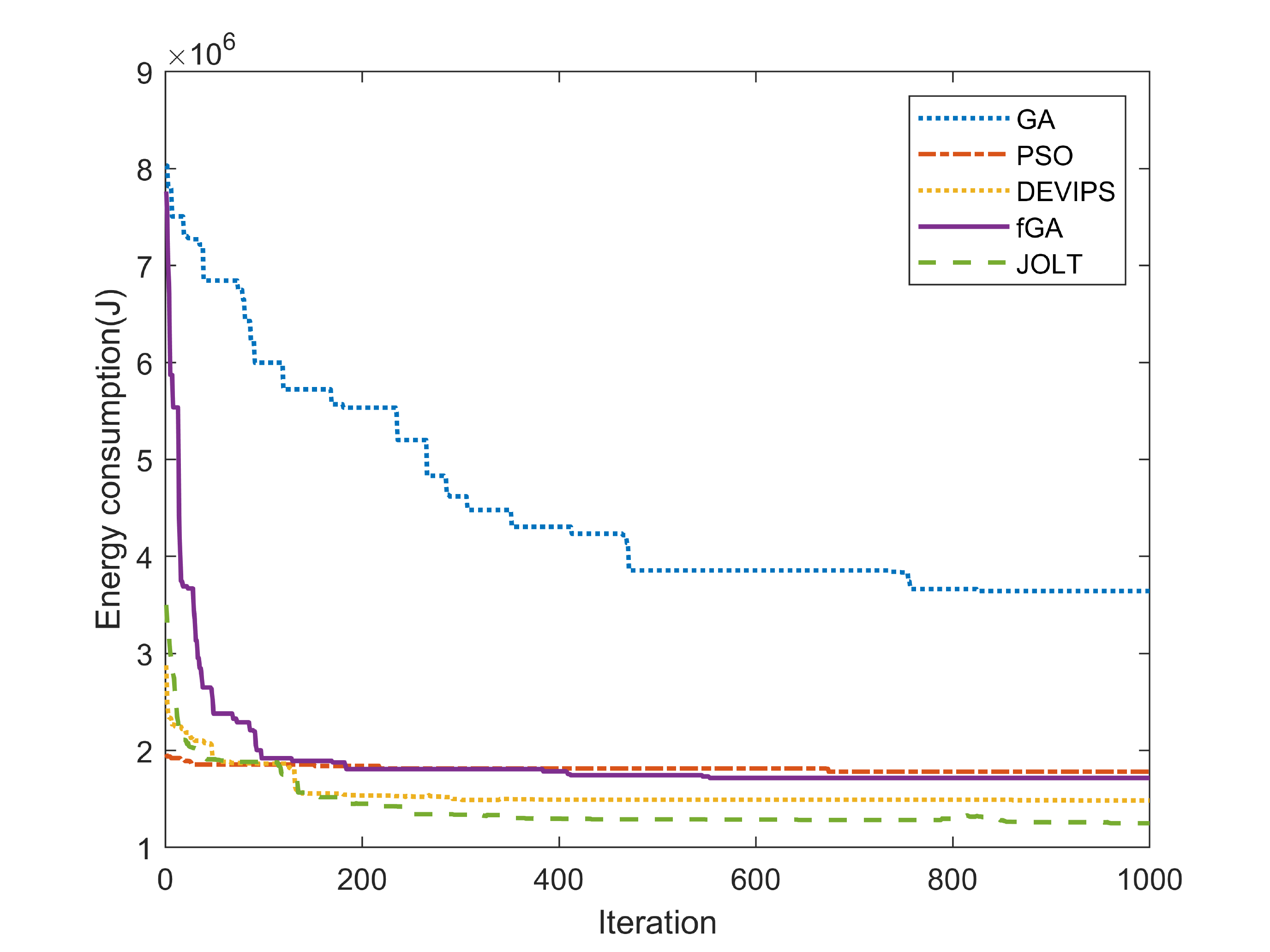}}
	\subfigure[Number of IoT devices is 300.]{\includegraphics[width=8.8cm]{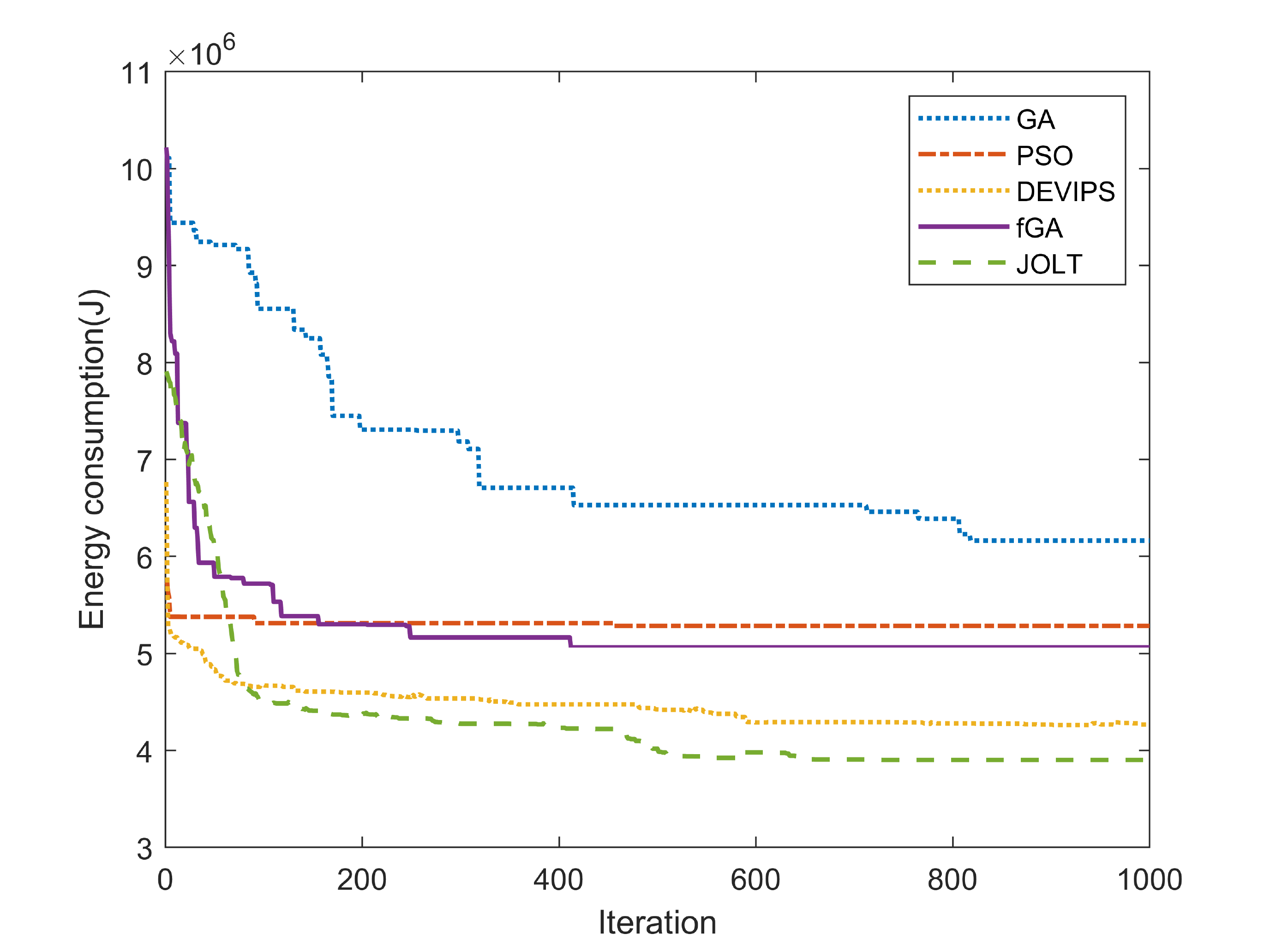}}
	\caption{Performance evaluation of different algorithms.}
	\label{fig:fig9}
\end{figure}
    \subsection{Performance evaluation of the JOLT framework}

    We evaluate the performance of the JOLT framework, DEVIPS\cite{8852863}, GA\cite{10.2307/24939139}, flexible genetic algorithm (fGA)\cite{ZHANG2017457} and PSO \cite{9275621} for joint optimization of UAV’s deployment and trajectory. Here are the parameter settings for different algorithms.
    	\begin{enumerate}
    		\item JOLT: The initial population size is set to 100; $b$ is set to 1; $\tau$ is set to 30; $\rho$ and $\mu$ are set to 1 and 4, respectively; $\beta$ is set to 0.1; $T_{\text {r}}$ is set to 5; $\zeta$ is set to 3.
    		
    		\item DEVIPS: The initial population size is set to 100; $F$ is set to 0.6;  $CR$ is set to 0.5. 
    		\item PSO: The population size is set to 10; Acceleration coefficients $\phi_{1}$ and $\phi_{2}$ are both set to 1.4; Inertia weight is set to 0.9.
    
    		\item GA: The population size is set to 10; The crossover probability $p_c$ is set to 0.9 and the mutation probability $p_m$ is set to 0.1 \cite{gao2018dendritic}.
    		\item fGA: The population size is set to 10; The length of the initial individual is set to 100; $p_c$ is set to 0.9 and $p_m$ is set to 0.1\cite{zhang2018online}.
    \end{enumerate}

We compare the energy consumption of the five algorithms in \textbf{Fig. \ref{fig:fig9}} with different number of IoT devices in the range of 1000 meters by 1000 meters. 
The locations of the initial stop points are generated in the same area randomly. It can be seen that the energy consumption of the JOLT framework in both figures are lower than other competitors. This is because the JOLT framework introduces AWOA to carry out the deployment optimization, which can find the optimal number of stop points and improve the exploration in search process. Moreover, it uses ERSOM to optimize the trajectory of the UAV, which improves efficiency of the trajectory planning.


In order to further evaluate the stability of the JOLT framework, three standard TSP data sets, i.e.,  att48, eil101 and tsp225 are selected as the benchmarks. \textbf{Table \ref{table_D}} shows the average energy consumption (Mean EC) and relative standard deviation (RStd) of the results under different learning rate $\beta$. It can be observed that different $\beta$ will affect the final results. The JOLT framework achieves the best result when $\beta=0.1$. Moreover, the RStd values of all experimental results are very low, which indicates the JOLT framework is stable and robust under different data sets.

\begin{table}[]
	\caption{Stability analysis of the JOLT framework.}
	\renewcommand{\arraystretch}{1.3}
	\setlength{\tabcolsep}{8.5mm}{
		\begin{tabular}{@{}llll@{}}
			\toprule
			Data set & $\beta$ & Mean EC& RStd \\ \midrule
			& 0.02 & 120.4106e4   & 0.099        \\
			att48  & 0.1  & 110.8519e4   & 0.058       \\
			& 0.5  & 121.9820e4   & 0.139        \\
			\cline{1-4}
			& 0.02 & 119.4103e4   & 0.111        \\
			eil101  & 0.1  & 109.4294e4   & 0.020        \\
			& 0.5  & 121.9820e4   & 0.151        \\
			\cline{1-4}
			& 0.02 & 116.0958e4   & 0.082         \\
			tsp225  & 0.1  & 109.1505e4   & 0.071        \\
			& 0.5  & 117.9440e4   & 0.101       \\ \bottomrule
	\end{tabular}}
	\label{table_D}
\end{table}  

\section{Conclusion}

In this paper, the JOLT framework has been proposed to optimize the deployment and trajectory in the UAV and IRS-Assisted IoT data collection system, with the objective of minimizing the energy consumption of the UAV and all IoT devices. The contributions can be summarized as follows: (1) A novel JOLT framework is designed to jointly optimize the deployment and trajectory of the UAV; (2) An AWOA method is proposed to find the optimal number of stop points and optimize the locations of stop points; (3) An ERSOM method is presented to learn the flight trajectory of the UAV by competitive learning among neurons.

For the experimental results, we firstly compare the AWOA with other WOAs in the deployment optimization. 
Then, the ERSOM is compared with RSOM and DEVIPS on three standard TSP data sets. 
Finally, the JOLT framework has been compared with different heuristic algorithms. the results have shown that the JOLT framework is stable and robust for different parameters and data sets, and it can achieve the best performance among the compared benchmarks.

For the application scope, the proposed JOLT framework can be applied in various areas, such as urban areas, where the communication links between UAV and users may be occasionally blocked by trees or buildings. 

For the future works, we plan to consider a multi-UAVs and IRSs-assisted IoT data collection system, where the deployment, trajectory and other piratical factors, like the battery limitation of the UAV can be jointly optimized in the real-world experiment.


\bibliographystyle{ieeetran}
\bibliography{bare_jrnl_bobo}
\begin{IEEEbiography}[{\includegraphics[width=1in,height=1.25in,keepaspectratio]{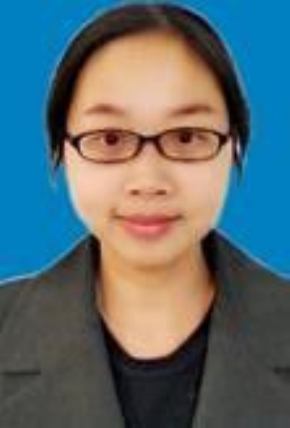}}]{Li Dong} received the B.S. and M.S. degrees in School of Physics and Electronics from Hunan Normal University, China, in 2004 and 2007, respectively. She received her Ph.D. degree in School of Geosciences and Info-physics from the Central South University, China, in 2018. Her research interests include machine learning, Internet of Things, and mobile edge computing.
\end{IEEEbiography}

\begin{IEEEbiography}[{\includegraphics[width=1in,height=1.25in,keepaspectratio]{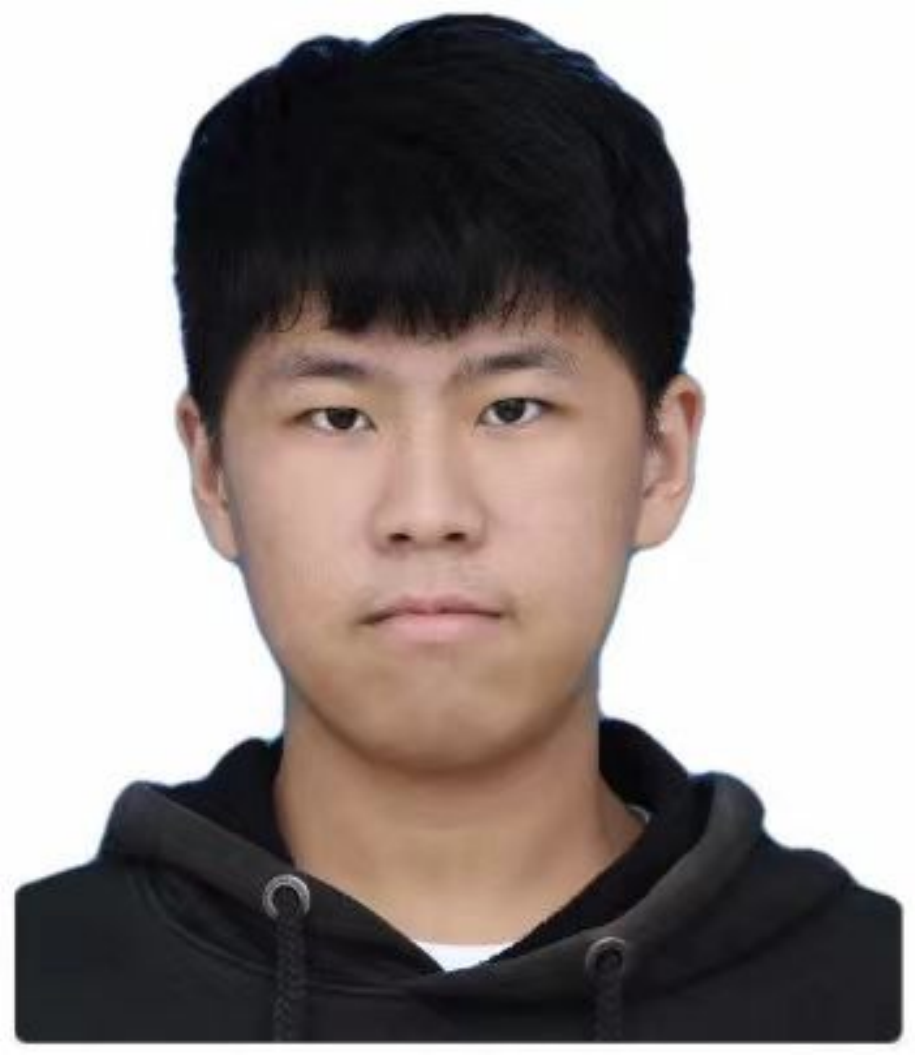}}]
{Zhibin Liu} received the B.S degree from Hunan Normal University, Changsha, China, in 2022. He is currently pursuing the master’s degree in Tianjin University, Tianjin, China. His main research interests include reinforcement learning and combinatorial optimization.
\end{IEEEbiography}

\begin{IEEEbiography}[{\includegraphics[width=1in,height=1.25in,keepaspectratio]{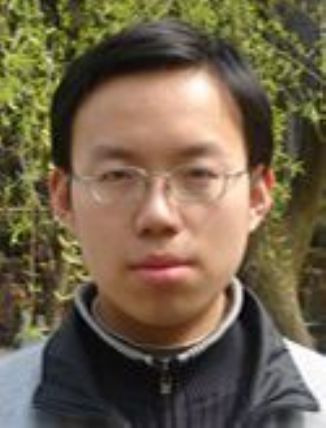}}]{Feibo Jiang} received his B.S. and M.S. degrees in School of Physics and Electronics from Hunan Normal University, China, in 2004 and 2007, respectively. He received his Ph.D. degree in School of Geosciences and Info-physics from the Central South University, China, in 2014. He is currently an associate professor at the Hunan Provincial Key Laboratory of Intelligent Computing and Language Information Processing, Hunan Normal University, China. His research interests include artificial intelligence,  Internet of Things, and mobile edge computing.
\end{IEEEbiography}

\begin{IEEEbiography}[{\includegraphics[width=1in,height=1.25in,keepaspectratio]{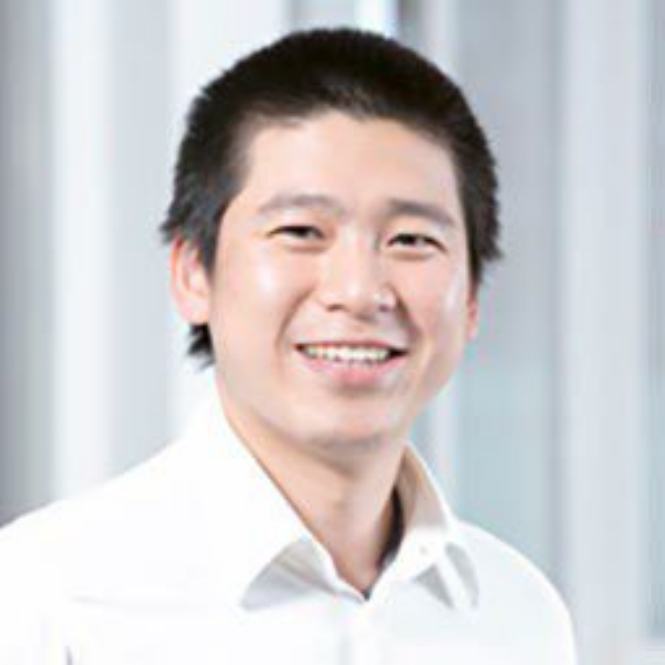}}]{Kezhi Wang}received his B.E. and M.E. degrees in School of Automation from Chongqing University, China, in 2008 and 2011, respectively. He received his Ph.D. degree in Engineering from the University of Warwick, U.K. in 2015. He was a Senior Research Officer in University of Essex, U.K. Currently he is a Senior Lecturer in Department of Computer and Information Sciences at Northumbria University, U.K. His research interests include wireless communication, mobile edge computing, UAV communication and machine learning.
\end{IEEEbiography} 

\newpage
\end{document}